\documentclass[letterpaper,journal]{IEEEtran}
\usepackage{amsmath,amsfonts}
\usepackage{algorithmic}
\usepackage{algorithm}
\usepackage{array}
\usepackage[caption=false,font=normalsize,labelfont=sf,textfont=sf]{subfig}
\usepackage{textcomp}
\usepackage{stfloats}
\usepackage{url}
\usepackage{verbatim}
\usepackage{graphicx}
\usepackage{cite}
\usepackage{microtype}
\usepackage{hhline}
\usepackage{makecell}
\usepackage{multirow}
\usepackage{xcolor}
\definecolor{lightred}{rgb}{1, 0.2, 0.2}
\definecolor{lightgreen}{rgb}{0.25, 0.75, 0.25}
\definecolor{iccvblue}{rgb}{0.26,0.39,0.85}
\usepackage{amssymb}
\usepackage{tabularx}
\usepackage{soul}
\usepackage{microtype}
\usepackage{graphicx}
\usepackage{booktabs}
\usepackage{colortbl}
\usepackage{pifont}

\usepackage[colorlinks=true]{hyperref}
\hypersetup{
    linkcolor=black, % 目录和交叉引用链接颜色
    urlcolor=black,  % 外部 URL 链接颜色
    citecolor=black, % 引用链接颜色
}

\definecolor{response}{rgb}{0.30,0.52,0.75}

\begin{document}

\title{GAPrompt++: Multi-Granular Geometry-Aware\\ Point Cloud Prompt for 3D Vision Model}

% \author{
% \thanks{\textsuperscript{*} Equal contribution}
% Zixiang Ai\thanks{Zixiang Ai is with Wangxuan Institute of Computer Technology, Peking University, Beijing, China. Email: zxAi25@stu.pku.edu.cn}\textsuperscript{*},
% Zhenyu Cui\thanks{Zhenyu Cui is with Wangxuan Institute of Computer Technology, Peking University, Beijing, China. Email: cuizhenyu@stu.pku.edu.cn}\textsuperscript{*},
% Yufei Guo\thanks{Yufei Guo is with Intelligent Science \& Technology Academy of CASIC, Beijing, China. Email: yfguo@pku.edu.cn}\textsuperscript{*},
% Wenwen Qiang\thanks{Wenwen Qiang is with Institute of Software, Chinese Academy of Sciences, Beijing, China. Email: qiangwenwen@iscas.ac.cn},
% Lei Chen\thanks{Lei Chen is with Department of Automation, Tsinghua University, Beijing, China. Email: leichenthu@tsinghua.edu.cn},
% Jiwen Lu (\textit{Fellow, IEEE})\thanks{Jiwen Lu is with Department of Automation, Tsinghua University, Beijing, China. Email: lujiwen@tsinghua.edu.cn},
% % Jiwen Lu \thanks{Jiwen Lu is with Department of Automation, Tsinghua University, Beijing, China. Email: lujiwen@tsinghua.edu.cn}\textsuperscript{\textit{Fellow, IEEE}},
% Jiahuan Zhou\thanks{Jiahuan Zhou is with Wangxuan Institute of Computer Technology, Peking University, Beijing, China. Email: jiahuanzhou@pku.edu.cn}\textsuperscript{$\dagger$}
% \thanks{\textsuperscript{$\dagger$} Corresponding author: Jiahuan Zhou}
% }

\author{
    Zixiang Ai\textsuperscript{*},
    Zhenyu Cui\textsuperscript{*},
    Yufei Guo\textsuperscript{*},
    Wenwen Qiang,
    Lei Chen,\\
    Jiwen Lu, \textit{Fellow, IEEE},
    Jiahuan Zhou
\thanks{* Equal contribution.}
\thanks{Zixiang Ai and Jiahuan Zhou are with the Wangxuan Institute of Computer Technology, Peking University, Beijing 100871, China.}
\thanks{Yufei Guo is with the Intelligent Science and Technology Academy of CASIC, Beijing 100041, China.}
\thanks{Wenwen Qiang is with the Institute of Software, Chinese Academy of Sciences, Beijing 100190, China.}
\thanks{Zhenyu Cui, Lei Chen, and Jiwen Lu are with the Department of Automation, Tsinghua University, Beijing 100084, China.}
\thanks{Corresponding author: Jiahuan Zhou (e-mail: jiahuanzhou@pku.edu.cn).}
\thanks{\copyright\ 2026 IEEE. Personal use of this material is permitted. Permission from IEEE must be obtained for all other uses, in any current or future media, including reprinting/republishing this material for advertising or promotional purposes, creating new collective works, for resale or redistribution to servers or lists, or reuse of any copyrighted component of this work in other works. Published version: \url{https://ieeexplore.ieee.org/document/11676080}. DOI: \url{https://doi.org/10.1109/TPAMI.2026.3729984}.}
}

\maketitle


% \IEEEpubid{0000--0000/00\$00.00~\copyright~2025 IEEE}

\begin{abstract}
Pre-trained 3D vision models have substantially advanced point cloud analysis, yet adapting them to downstream tasks via full fine-tuning is computationally expensive and storage-intensive. Parameter-Efficient Fine-Tuning (PEFT) offers a promising alternative by reducing both adaptation cost and storage burden. However, existing prompting-based approaches ignore the intrinsic geometric structures of point clouds, thereby limiting their adaptation capability. This limitation stems from their inability to encode both fine-grained geometric cues and coarse-grained structural semantics, as well as failing to propagate such information effectively through the model hierarchy. 
To address these challenges, we propose GAPrompt++, a multi-granular geometry-aware prompting method that provides richer geometric guidance for efficient 3D task adaptation. Specifically, we introduce a Point Shift Prompter that extracts multi-granular geometric features across different scales, enabling instance-specific geometric adjustments during adaptation. Next, a Keypoint Prompter adaptively generates point-level prompts to highlight local geometric saliency and fine-grained structural details. Furthermore, a Prompt Propagation mechanism injects these multi-granular geometric cues throughout the feature extraction hierarchy, strengthening the ability to capture essential geometric characteristics. 
Extensive experiments show that GAPrompt++ achieves state-of-the-art performance among prompting-based PEFT methods and even surpasses full fine-tuning across diverse benchmarks, while requiring less than 2\% trainable parameters. In addition, to address the saturation of existing evaluation datasets, we construct two more challenging benchmarks derived from 3D Gaussian Splatting and Multi-View Stereo reconstruction, offering diverse and realistic point cloud scenarios to promote future research. Our code will be released at $\texttt{https://github.com/PKU-OV3-LAB/GAPromptPlus.git}$.

% Our code is available at \url{https://github.com/zhoujiahuan1991/ICML2025-GAPrompt}. 
\end{abstract}


\section{Introduction}
% \IEEEPARstart{T}{he} advent of scanning sensor devices has significantly facilitated the acquisition of 3D point cloud data, an inherently irregular and unstructured geometric representation. This advancement has propelled the development of various 3D vision applications, including 3D reconstruction~\cite{xu2022point, Lu_2024_CVPR} and autonomous driving~\cite{zhao2024pnerfloc}. Recently, pre-trained 3D vision models~\cite{yu2022PointBERT, zhang2022PointM2AE,zha2023PointFEMAE} have shown remarkable performance in processing point cloud data, enabling their direct application to a variety of downstream 3D tasks through full fine-tuning. However, fine-tuning the entire pre-trained model incurs substantial computational costs and necessitates a large quantity of labeled data. Moreover, without freezing the pre-trained model, there exists a considerable risk of catastrophic forgetting, potentially resulting in the loss of critical pre-trained knowledge.

\IEEEPARstart{P}{oint} cloud data acquisition has been greatly advanced by modern sensing technologies such as LiDAR, structured-light scanners, and RGB-D cameras, which provide sparse or dense point clouds as a flexible geometric representation of real-world scenes. Meanwhile, the rapid development of 3D reconstruction pipelines, including Multi-View Stereo (MVS)~\cite{mvsnet}, Neural Radiance Fields (NeRF)~\cite{nerf}, and 3D Gaussian Splatting~\cite{gaussiansplatting}, has further reduced the cost of generating large-scale 3D assets.

% \begin{figure}[!t]
% \centering
% \includegraphics[width=0.9\linewidth]{images/header.pdf}
% \caption{Our GAPrompt++ compares to full fine-tuning and existing prompting methods. We compare the classification accuracy on the hardest variant of \textit{ScanObjectNN}\cite{scanobjectnn} based on pre-trained Point-PQAE~\cite{pointpqae}.}
% \label{fig-header}
% % \vspace{-10 pt}
% \end{figure}

\begin{figure}[!t]
\centering
\includegraphics[width=\linewidth]{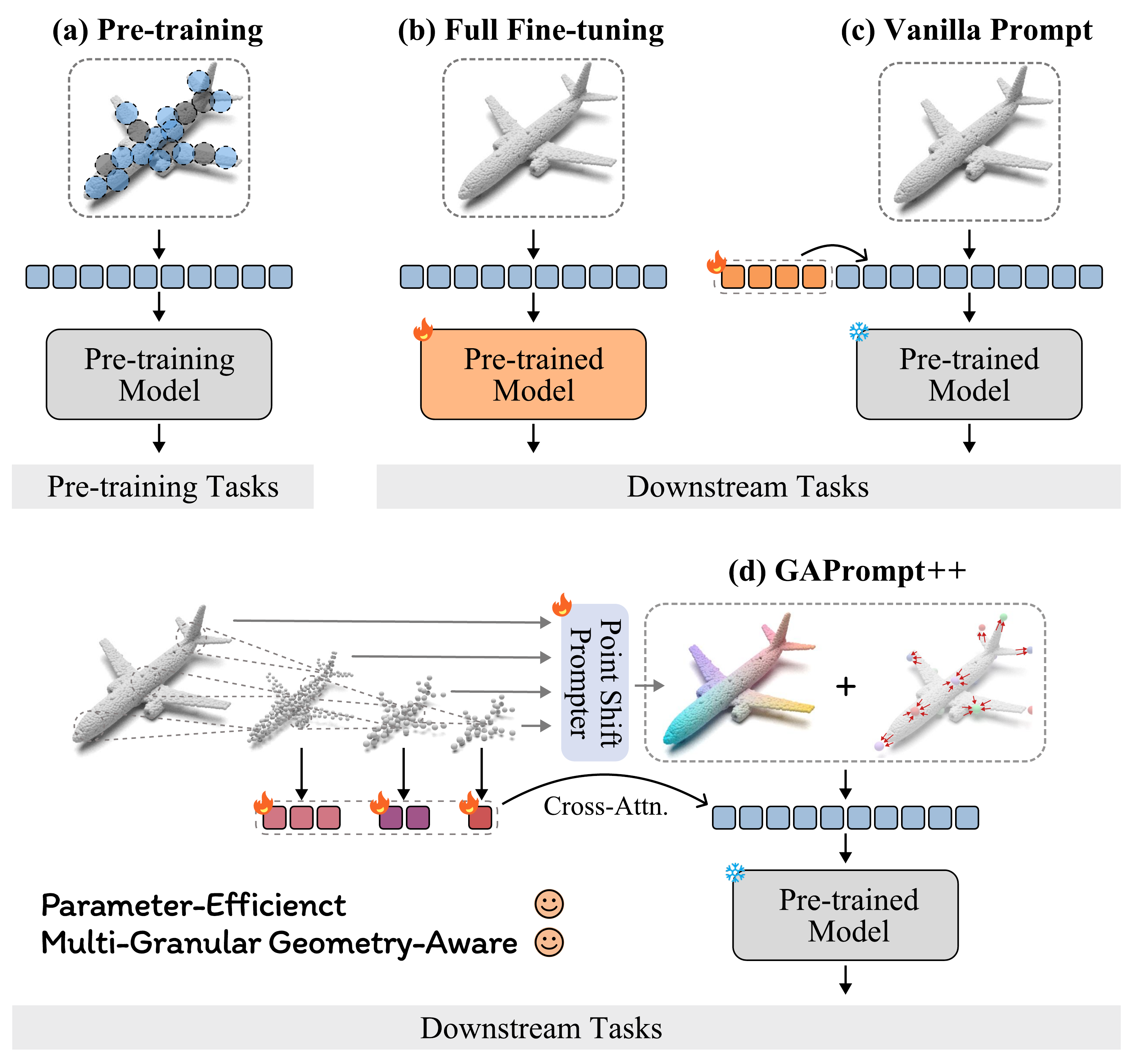}
\vspace{-15 pt}
\caption{Methods for adapting (a) pre-trained 3D vision models. (b) Fine-tuning updates the entire model parameters. (c) Vanilla prompt-based methods adapt the model pre-trained models by reformulating the input at the token level. (d) Our GAPrompt++ introduces explicit keypoint prompts and geometry-enhanced token prompts with instance-specific multi-granular features, extracted by the Point Shift Prompter.}
\label{fig-motivation-header}
\vspace{-15 pt}
\end{figure}

Attributed to large-scale datasets, pre-training 3D vision models~\cite{yu2022PointBERT, zhang2022PointM2AE, zha2023PointFEMAE} have demonstrated impressive generalization ability across diverse point cloud tasks. By scaling both model parameters and pre-training datasets, these models provide powerful geometric priors that can be transferred to downstream tasks through full fine-tuning. Despite their strong performance, full fine-tuning remains computationally expensive, as it requires updating all parameters of the pre-trained backbone and storing a separate model instance for each downstream task. Moreover, adequately adapting such large-capacity models typically demands substantial amounts of annotated 3D data, and naively fine-tuning the full model may disrupt its pre-trained knowledge, causing catastrophic forgetting and degraded generalization.

To tackle the above challenges, Parameter-Efficient Fine-Tuning (PEFT) has gained traction in natural language processing and 2D vision due to its ability to adapt large pre-trained models using lightweight trainable modules while keeping the backbone frozen, including prompt tuning methods~\cite{lester2021power, li2021prefix, liu2024insvp} and adapter tuning methods~\cite{adapter, prompt}, both of which aim to bridge task-specific gaps with efficient backbone reuse. When applied to 3D point clouds with sparsity, irregular sampling, and permutation invariance, existing PEFT methods still face substantial challenges. Specifically, the widely adopted random initialization strategy~\cite{vpt} in prompt learning often struggles to align with point distributions, causing unstable optimization in downstream tasks. Besides, existing adapter-based approaches capture fine-grained geometric structures encoded by spatial point arrangements only partially, due to their reliance on coarse-grained latent features. As a result, existing PEFT methods in 3D vision leave substantial room for stronger use of the most discriminative geometric cues, which constrains their adaptation capability and overall performance.

% \begin{figure*}[!t]
%     \centering
%     \includegraphics[width=\textwidth]{images/motivation.pdf}
%     \caption{Methods for adapting (a) pre-trained 3D vision models. (b) Fine-tuning updates the entire model parameters. (c) Vanilla prompt-based methods adapt the model pre-trained models by reformulating the input at the token level. (d) Our GAPrompt++ introduces explicit keypoint prompts and geometry-enhanced token prompts with instance-specific multi-granular features, extracted by the Point Shift Prompter.}
%     \label{fig-motivation}
%     % \vspace{-8pt}
% \end{figure*}

\begin{figure}[!t]
\centering
\includegraphics[width=\linewidth]{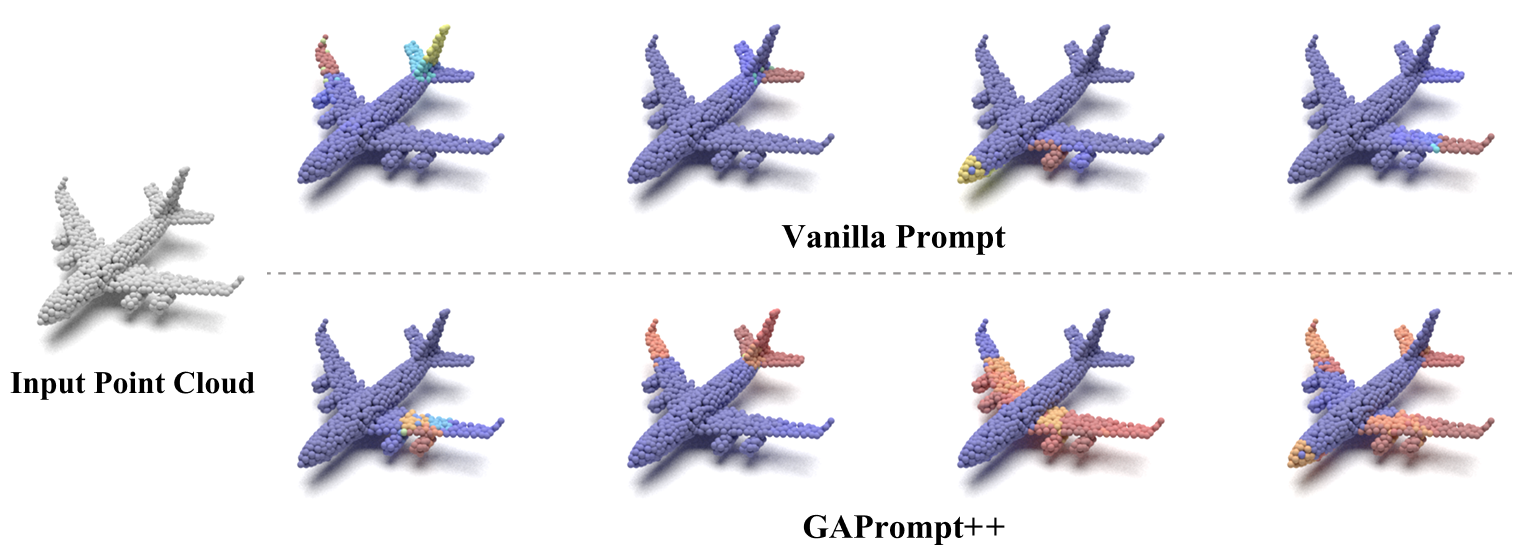}
\vspace{-20 pt}
\caption{Visualization of feature correspondence between prompt tokens and input tokens, where warmer colors indicate higher values.. Vanilla prompt learning mainly forms localized correspondences confined to individual token scales. In contrast, GAPrompt++ captures multi-granular geometric cues, enabling correspondences to extend across tokens and align with coherent structural components of the 3D shape.}
\label{fig-effect-header}
\vspace{-15 pt}
\end{figure}

Recently, several studies~\cite{idpt, zhou2024DAPT} have recognized the critical need for specifically designed 3D PEFT methods and have scratched out the initial steps. These approaches generally construct auxiliary networks to model token interactions and dynamically generate token-level prompts, or employ adaptive adapters that jointly refine input tokens and synthesize prompt tokens. However, as shown in Fig.~\ref{fig-effect-header}, these methods usually concentrate primarily on encoded input tokens, which correspond to local patches at a single scale, failing to capture the intrinsic geometric structures of point clouds, which severely limits their ability to achieve competitive performance. Furthermore, the derived fine-grained geometric cues and coarse-grained structural semantics are more difficult to capture and exploit due to the limitation of propagating multi-granular information effectively through the single-scale modeling paradigm.

To this end, we propose GAPrompt++, a multi-granular geometry-aware prompting framework that provides richer geometric guidance for efficient 3D task adaptation. First, we introduce a Point Shift Prompter that extracts geometric features across multiple scales, enabling instance-specific geometric adjustments that better align pre-trained features with downstream objectives. Second, a Keypoint Prompter adaptively generates point-level prompts to emphasize local geometric saliency and fine-grained structural cues. Additionally, a Prompt Propagation mechanism spreads these multi-granular geometric signals throughout the hierarchical feature extraction process, strengthening the model’s ability to capture essential geometric characteristics.

Beyond purely 3D backbones, the growing convergence of 3D vision with natural language and 2D image understanding underscores the need for PEFT approaches capable of supporting multi-modal adaptation. A notable finding is that our geometry-aware prompting design is inherently modality-agnostic: GAPrompt++ effectively adapts large-scale text encoders (e.g., CLIP) and vision encoders (e.g., DINOv3) to 3D downstream tasks, despite their heterogeneous input structures. This reveals that geometric cues extracted directly from point clouds can function as a general adaptation prior across modalities, substantially broadening the applicability of GAPrompt++ in unified multi-modal 3D learning.

In summary, key contributions of this work are four-folded:
\begin{itemize}
\item{We propose GAPrompt++, a multi-granular geometry-aware prompting method tailored for 3D vision models, achieving performance matching or surpassing full fine-tuning while substantially reducing computational and storage overhead.}
\item{We introduce three core components, Point Shift Prompter, Keypoint Prompter, and Prompt Propagation, which collectively enable the model to capture and exploit multi-granular geometric information inherent in point clouds, enhancing its representational capability.}
% \item{Extensive experiments across diverse benchmarks demonstrate that GAPrompt++ delivers state-of-the-art performance among PEFT methods and consistently outperforms full fine-tuning under stringent parameter budgets.}
\item{To address the saturation of existing evaluation datasets, we construct two new, more challenging benchmarks derived from 3D Gaussian Splatting and Multi-View Stereo reconstruction, providing diverse and realistic point cloud scenarios to foster future research.}
\item{We demonstrate that GAPrompt++ naturally extends to multi-modal adaptation, enabling effective transfer of CLIP text encoders and DINOv3 vision encoders to 3D tasks, revealing geometry as a unifying adaptation interface across modalities.}
\end{itemize}

% This manuscript is an extension of our previous conference paper~\cite{ai2025gaprompt}. In this study, we have made extensive extensions including (1) To mitigate the interference of domain shifts during historical prototype utilization, a privacy-friendly inputlevel alignment mechanism is developed to bridge the gap between the new and old domains. (2) To fully leverage the new and aligned data, feature-level constraints are introduced to form a dual-side alignment scheme, enhancing the model’s capacity to acquire new knowledge and retain historical knowledge. (3) Comprehensive theoretical and experimental analyses are provided in Sec.~\ref{analysis} to validate the effectiveness of our algorithm compared to existing works. (4)Five additional LReID dataset orders beyond the two used in the original study are introduced, demonstrating the robustness of our method under varied domain shift scenarios. (5) More ablation studies and detailed analyses, including learning trends, ablation studies on module effectiveness, adaptability to the vision foundation model, and visualization results, are presented in Sec. V.

This manuscript is an extension of our previous conference paper~\cite{ai2025gaprompt} with several significant advances. 
(1) We overcome the limitations of existing prompt learning approaches in capturing fine-grained local geometry and global hierarchical semantics in point clouds.
(2) We introduce two new datasets, \textit{GSModel60} and \textit{uCO3D80}, to alleviate benchmark saturation and better align with real-world 3D reconstruction pipelines. 
(3) We show that GAPrompt++ can effectively adapt pre-trained models from other modalities, such as text and 2D images, thereby reducing reliance on large point cloud data. 
{(4) We provide an optimal-transport-inspired analytical perspective and discussion to help interpret the behavior of the proposed multi-granular prompting design.}
(5) We deliver richer empirical evidence, including ablations, multi-modal comparisons, and evaluations on more challenging settings, further validating the effectiveness and generality of our approach.

\section{Related Work}
\subsection{Point Cloud Representation Learning}
Point cloud representation is fundamental to a wide spectrum of 3D applications, including autonomous driving~\cite{rt2,openvla}, robotics~\cite{asap}, and embodied intelligence systems~\cite{navgpt,objectnav}. Early foundational works~\cite{pointnet,pointnet++,dgcnn,pointnext} demonstrated that unordered and sparse point sets can be effectively processed using MLP-based or local aggregation architectures, laying the groundwork for modern 3D understanding. The emergence of Transformer-based and state-space models (e.g., Vision Transformer~\cite{dosovitskiy2020vit}, Vision Mamba~\cite{visionmamba}) has further advanced 3D representation learning by enabling global receptive fields and scalable modeling capacity.

With increasing model capacity and the availability of large-scale 3D datasets, self-supervised pre-training has become the dominant paradigm for learning generalizable 3D representations. Existing methods mainly fall into two categories: contrastive learning and masked modeling.
Contrastive pre-training methods~\cite{pointcontrast} acquire geometric-aware representations by distinguishing structurally similar samples from dissimilar ones, typically leveraging invariances to rotation, perturbation, and spatial transformation. Beyond purely geometric pre-training, cross-modal contrastive approaches~\cite{clip2point,pointclipv2,ACT_rotate,qi2023ReCon,recon++} align point clouds with pretrained language or image models~\cite{clip,dinov3}, enabling the transfer of rich semantic priors from other modalities.
Inspired by masked modeling in natural language and vision~\cite{bert,mae}, masked point modeling approaches~\cite{yu2022PointBERT,mae,maskpoint} reconstruct randomly occluded local regions to learn geometric completion priors. Subsequent works, including Point-M2AE~\cite{zhang2022PointM2AE}, Point-FEMAE~\cite{zha2023PointFEMAE}, PCP-MAE~\cite{pcpmae}, and Point-PQAE~\cite{pointpqae}, explored hierarchical masking strategies, improved decoders, and richer geometric objectives. PointGPT~\cite{pointgpt} reformulated the reconstruction task as sequential regression and curated large hybrid datasets for post-pretraining, achieving notable downstream gains. Another line of work~\cite{tap,i2pmae,pointsd} projects point clouds onto 2D planes and utilizes strong 2D priors from models such as DINO~\cite{dino} or diffusion models~\cite{diffusion}, highlighting the benefit of cross-dimensional knowledge transfer. Attributed to geometric-rich representations obtained from pre-training, fine-tuning these models consistently outperforms classical supervised architectures across a wide range of downstream tasks.

Despite these advances, full fine-tuning remains computationally expensive. Scaling backbone capacity and expanding downstream tasks require maintaining task-specific replicas of the full model, which is increasingly impractical for large 3D architectures and multi-task deployment. These limitations motivate the need for parameter-efficient fine-tuning (PEFT) tailored for 3D. Our method captures both global shape structure and local topological variations and injects them into the backbone through lightweight prompt tokens, providing an effective and scalable alternative to full fine-tuning.

% The overall pipeline of GAPrompt++. Given an input point cloud, the Point Shift Prompter captures multi-granular geometric features and applies fine-grained shifts to each point. The Keypoint Prompter then identifies salient local structures and generates instance-specific keypoints as prompts. During feature extraction, learnable prompt tokens are prepended and fused with the extracted multi-granular geometric features. The concatenated tokens are subsequently passed through Prompt Propagation, which injects intrinsic geometric cues throughout the model hierarchy. Finally, the propagated tokens are passed into the pre-trained backbone to generate downstream task predictions.

\subsection{Parameter-Efficient Fine-Tuning}
With the rapid growth of pre-training models in capacity, fully fine-tuning all parameters for downstream tasks has become increasingly expensive. To alleviate this burden, researchers in NLP and 2D vision have developed a variety of Parameter-Efficient Fine-Tuning (PEFT) techniques. Among them, prompt tuning~\cite{li2021prefix,vpt} introduces learnable latent tokens to steer pretrained models while keeping backbone parameters frozen, while adapter-based methods~\cite{adapter} insert lightweight modules into intermediate layers to refine latent feature distributions. Numerous extensions~\cite{vp,dept,adaptformer,biadaptformer} demonstrate that such designs can match full fine-tuning performance with only a small number of trainable parameters.

\begin{table}[!t]
\centering
\caption{Comparisons of our proposed two challenging datasets with existing benchmarks.}
\setlength{\tabcolsep}{3pt}
\resizebox{\linewidth}{!}{
\begin{tabular}{lccccc}
\toprule
\textbf{Dataset}&\textbf{Sample}&\textbf{Category}&\textbf{Real World}&\textbf{Reconstruction}&\textbf{Data Source}\\ 
\midrule
\textit{ShapeNet55} & 16,880 & 55 & - & - & CAD Model Sampling \\ 
\textit{ModelNet40} & 12,311 & 40 & - & - & CAD Model Sampling  \\
\textit{ShapeNetPart} & 15,008 & 16 & - & - & CAD Model Sampling  \\
\textit{ScanObjectNN} & 14,298 & 15 & $\checkmark$ & - & RGB-D Scene Scan \\
\midrule
\textit{GSModel60} & 13,847 & 60 & $\checkmark$ & $\checkmark$ & 3D Gaussian Splatting \\
\textit{uCO3D80} & 20,464 & 80 & $\checkmark$ & $\checkmark$ & Multi-View Stereo \\
\bottomrule
\end{tabular}
}
\label{tab-dataset-comparison}
% \vspace{-6 pt}
\end{table}

\begin{figure}[!t]
\centering
\includegraphics[width=\linewidth]{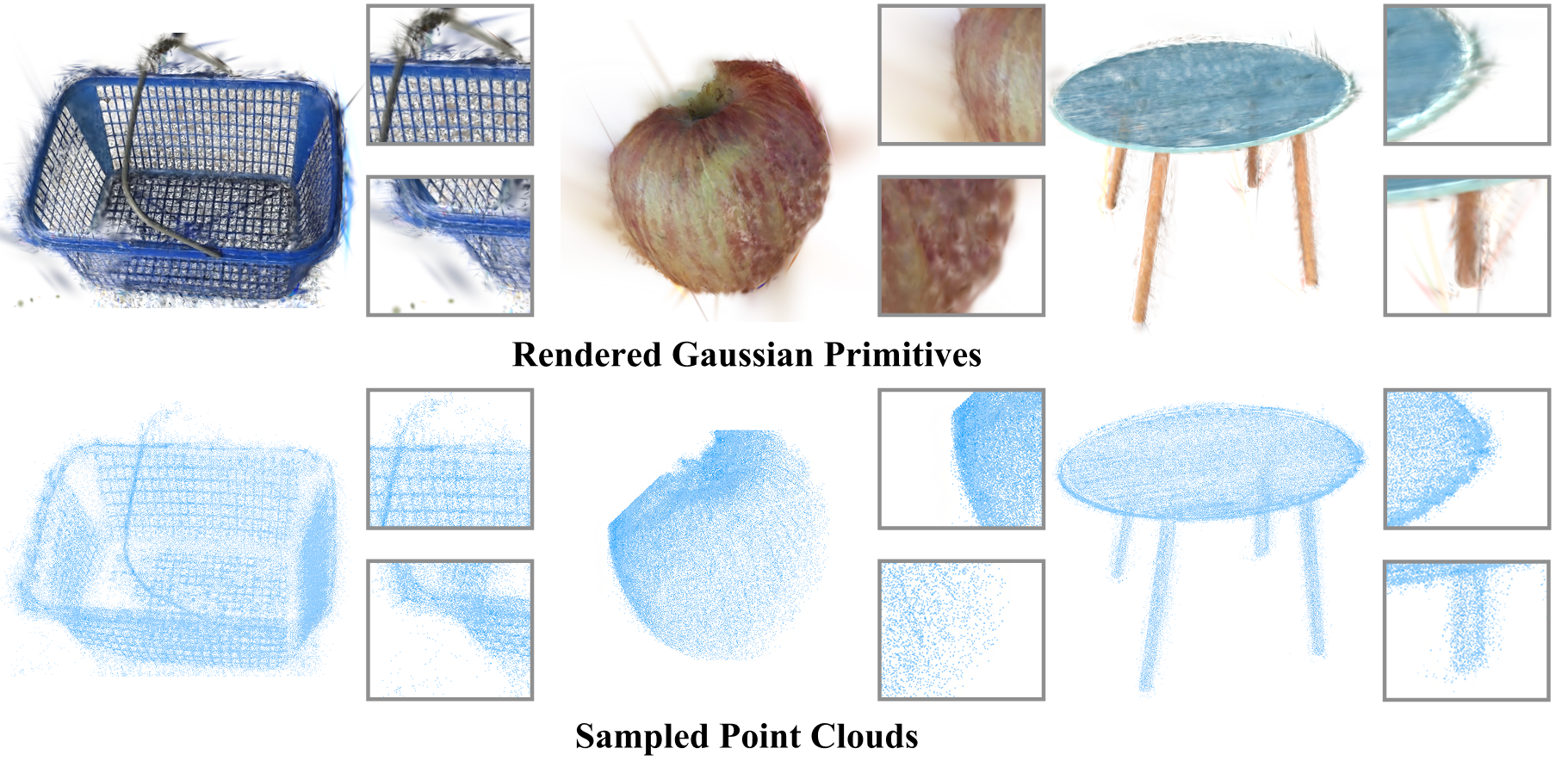}
\vspace{-18pt}
\caption{
Examples from the \textit{GSModel60} dataset. Top: rendered Gaussian primitives. Bottom: point clouds sampled from the reconstructed primitives.
}
\label{fig-gsmodel60}
\end{figure}

\begin{figure}[!t]
\centering
\includegraphics[width=\linewidth]{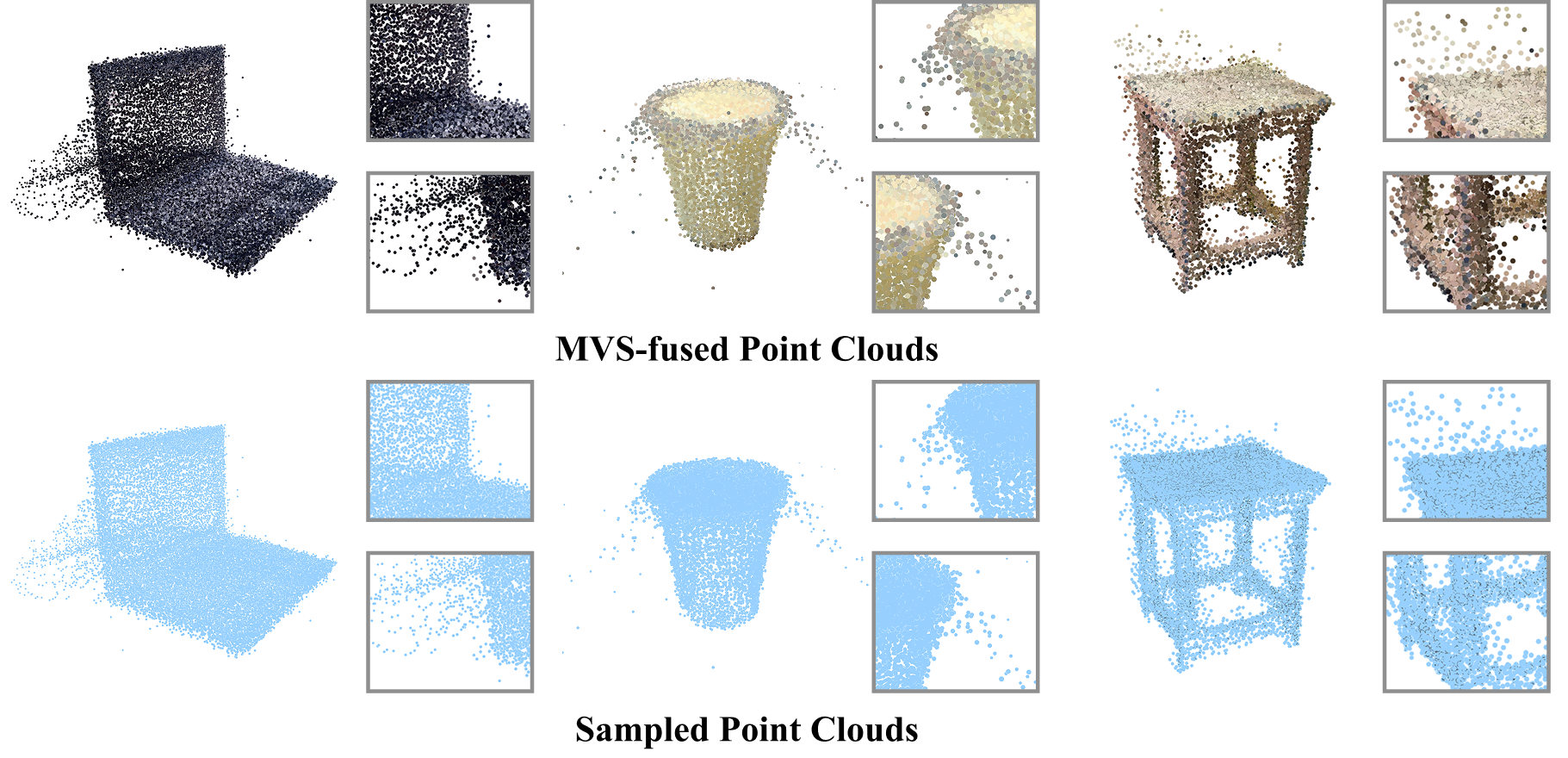}
\vspace{-18pt}
\caption{
Examples from the \textit{uCO3D80} dataset. Top: MVS-fused dense colored point clouds. Bottom: underlying point clouds sampled for analysis.}
\vspace{-10 pt}
\label{fig-uco3d80}
\end{figure}

However, directly transferring PEFT paradigms to 3D vision is challenging due to the sparsity, irregularity, and permutation invariance of point clouds, which fundamentally differ from the grid-structured data in text and images. Recent efforts have therefore explored 3D-specific PEFT, which can be broadly categorized into reparameterization-based, adapter-based, and prompt-based approaches.
Reparameterization methods such as PointLoRA~\cite{wang2025pointlora} and MoST~\cite{han2025most} adopt low-rank adaptation combined with multi-scale token selection or reformulate dense update matrices using sparse monarch structures to reduce parameter overhead.
Adapter-based methods function by tailoring architectural modules to 3D representations. DAPT~\cite{zhou2024DAPT} introduces dynamic scale adapters for joint feature refinement and prompt generation, while PointGST~\cite{liang2025pointgst} decomposes latent representations into the graph spectral domain to enable flexible adaptation to geometric variations. PMA~\cite{zha2025pma} further enhances by injecting mamba blocks that capture complementary contextual dependencies from intermediate layers. {{GEM~\cite{gem} specifically designs spatial and context adapters for semantic  segmentation tasks.}} 
Prompt-based methods offer another perspective by modifying the input token space. IDPT~\cite{idpt} dynamically generates prompts via an EdgeConv network to model local interactions, achieving promising performance but incurring notable computational cost. Point-PEFT~\cite{tang2024Point-PEFT} constructs a prompt bank from pre-processed training features to introduce data-driven priors, whereas PPT~\cite{zhang2024ppt} emphasizes the importance of positional cues by designing explicit positional prompts, albeit neglecting the local geometric structures and at the cost of significantly increasing token numbers.

Despite these advances, existing 3D PEFT methods often fail to jointly capture fine-grained local geometry and global semantic context, limiting their adaptability across diverse tasks. To address this limitation, we propose GAPrompt++, a multi-granular geometry-aware prompting framework that integrates both local structural cues and global semantic priors to provide richer geometric guidance for efficient model adaptation.

\subsection{Point Cloud Analysis Benchmarks}
The rapid development of depth sensors~\cite{scannet} and 3D reconstruction techniques~\cite{mvsnet,gaussiansplatting} has substantially reduced the cost of acquiring point cloud data, giving rise to a series of large-scale 3D datasets. ShapeNet~\cite{shapenet55} has long served as a foundational pre-training dataset, containing over 3 million textured CAD models, including the widely adopted ShapeNet-Core subset of 52K curated meshes with high-quality annotations. ModelNet~\cite{modelnet40} provides 12K CAD models across common household categories, while OmniObject3D~\cite{omniobject3d} extends object diversity with 6K instances. More recent datasets, such as Objaverse~\cite{objaverse} and Objaverse-XL~\cite{objaversexl}, further scale to 800K and 10.2M assets, enriched with textual descriptions to support large-scale 3D representation learning and multimodal understanding.

\begin{figure*}[t]
\vspace{-8pt}
\centering
\includegraphics[width=\textwidth]{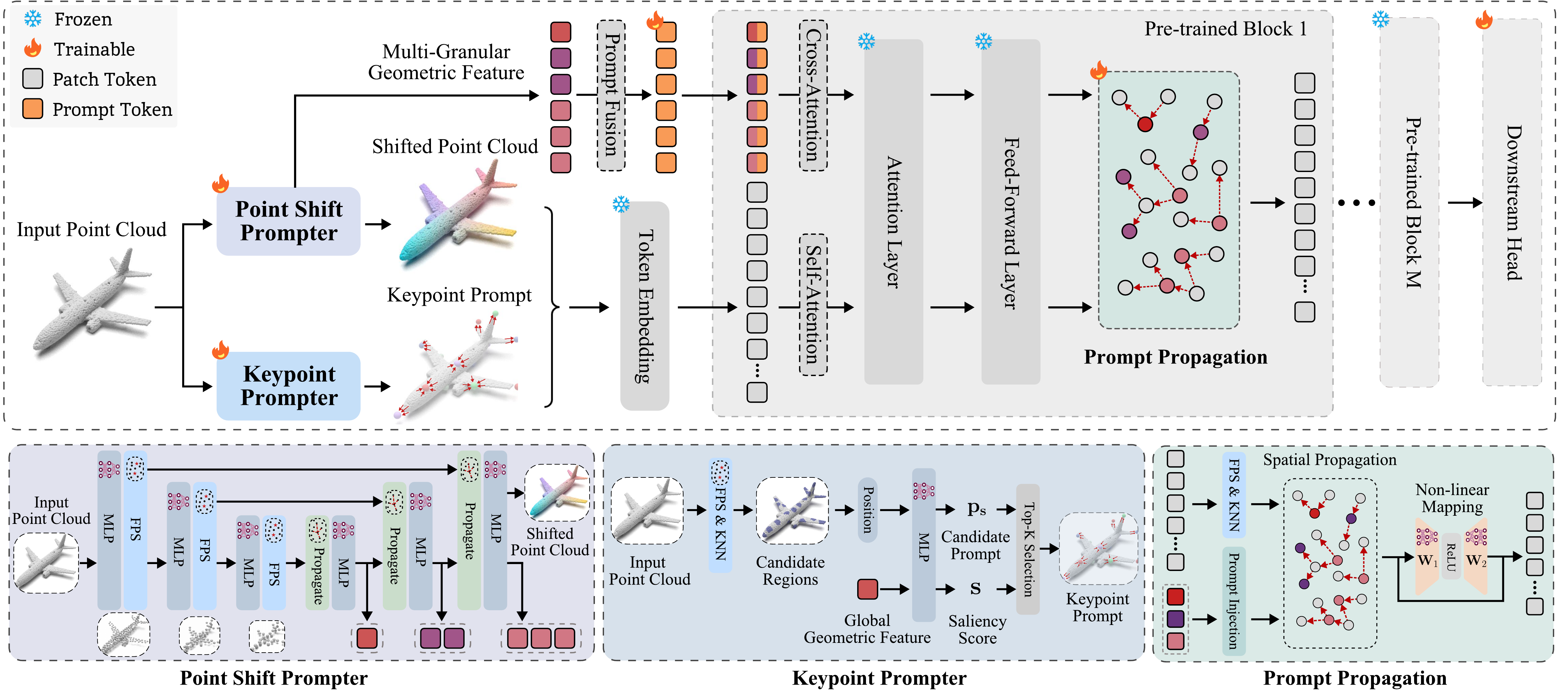}
\caption{
The overall pipeline of GAPrompt++. Given an input point cloud, the Point Shift Prompter extracts multi-granular geometric features and performs fine-grained point shifts, while the Keypoint Prompter identifies salient local structures and generates instance-specific keypoint prompts. During feature extraction, learnable prompt tokens are prepended and fused with multi-granular geometric features and then are processed by the Prompt Propagation, which injects geometric cues across the model hierarchy. Finally, the propagated tokens are passed through the pre-trained backbone to make downstream predictions.
}
\vspace{-12 pt}
\label{fig-pipeline}
\end{figure*}

Although large-scale 3D data has substantially advanced pre-training, commonly used downstream benchmarks, especially for classification, are nearly saturated. ModelNet40~\cite{modelnet40} contains clean, uniformly sampled CAD models that lack real-world sensing complexity, while ScanObjectNN~\cite{scanobjectnn}, despite introducing clutter and occlusion, remains limited in category diversity and scene variability. 
Recent state-of-the-art methods~\cite{pointgpt,recon++} already surpass 97\% accuracy on ModelNet40 and 99\% on ScanObjectNN, leaving minimal headroom to demonstrate meaningful progress. These saturation issues reflect two core limitations: the idealized noise-free acquisition setup of CAD datasets, and the insufficient alignment with modern reconstruction pipelines such as Multi-View Stereo (MVS)~\cite{mvsnet} and 3D Gaussian Splatting (3DGS)~\cite{gaussiansplatting}, which produce inherently noisy and irregular point distributions.

To better reflect contemporary 3D data characteristics and support more discriminative evaluation, we construct two new challenging benchmarks, \textit{GSModel60} and \textit{uCO3D80}, summarized in Table~\ref{tab-dataset-comparison}. They are derived respectively from 3D Gaussian Splatting reconstruction and unconstrained multi-view reconstruction, introducing richer geometric variability, realistic reconstruction artifacts, and broader semantic coverage, as illustrated in Fig.~\ref{fig-gsmodel60} and Fig.~\ref{fig-uco3d80}.

\section{The Proposed Method} 
We present GAPrompt++, an efficient fine-tuning method for pre-trained 3D vision models. It consists of three geometry-aware prompt modules: the Point Shift Prompter, the Keypoint Prompter, and the Prompt Propagation mechanism. As shown in Fig.~\ref{fig-pipeline}, given a pre-trained 3D transformer with $M$ blocks and a downstream task, we freeze the backbone and solely update newly introduced modules together with the task head.

\subsection{Preliminary for Pre-trained Point Cloud Model}
Transformer-based architectures~\cite{dosovitskiy2020vit} have become the dominant backbone for point cloud pre-training due to their strong capability in modeling long-range dependencies. Given an input point cloud $\mathbf{x} \in \mathbb{R}^{N\times3}$, the points are first partitioned into local patches using farthest point sampling (FPS) and k-nearest neighbors (KNN). Each patch is then embedded into a token sequence $\mathbf{h}_0 \in \mathbb{R}^{L{\times}D}$, where $L$ and $D$ denote the token length and embedding dimension.

A pre-trained 3D Transformer typically contains $n$ hierarchical blocks, each composed of a self-attention module and a feed-forward network (FFN). The input tokens are progressively updated as:
\begin{equation}
\mathbf{\bar{h}}_i = \texttt{Attn}(\mathbf{h}_i),
\label{attn}
\end{equation}
\begin{equation}
\mathbf{h}_{i+1} = \texttt{FFN}(\mathbf{\bar{h}}_i),
\label{ffn}
\end{equation}
where $\mathbf{h}_{i+1} \in \mathbb{R}^{L{\times}D}$ represents output features of the $i$-th block. After that, these high-level representations are subsequently consumed by downstream heads for tasks such as classification or segmentation.

Notably, the attention and FFN layers dominate the model's parameter and computational cost. Within GAPrompt++, all backbone parameters are frozen, and only the proposed geometry-aware prompting modules and the downstream prediction head remain trainable.

\subsection{Point Shift Prompter}
To capture both fine-grained geometric details and coarse structural semantics, we design a Point Shift Prompter that extracts multi-granular geometric features from the input point cloud. The fine-grained features are further processed by a shallow shift head to predict a coordinate shift for each point, enabling point-level geometric modulation.

Specifically, to acquire geometric features of point clouds with diverse granularities, we gradually process the points in a hierarchical manner. As shown in Fig.~\ref{fig-pipeline}, the input point cloud $\mathbf{x}_0 \in \mathbb{R}^{N\times3}$ coordinates are encoded into spatial features $\mathbf{f}_0$ with a tiny pointnet\cite{pointnet} layer, denoted as $\mathbf{\Phi}$. We then downsample $\mathbf{x}_0$ via farthest point sampling (FPS) to obtain $\mathbf{x}_1 \in \mathbb{R}^{\frac{N}{2}{\times}3}$, and iteratively extract higher-level semantics from progressively coarser point sets:
\begin{equation} 
    \mathbf{f}_{i+1} = \mathbf{\Phi}(\mathbf{f}_{i}) ,
\end{equation}
\begin{equation} 
    \mathbf{x}_{i+1} = \texttt{FPS}(\mathbf{x}_{i}),
\end{equation}
where ${0 \le i \le m-1}$. After $m$ layers, we obtain hierarchical point sets $\{\mathbf{x}_0,\ldots,\mathbf{x}_m\}$ and corresponding feature sets $\{\mathbf{f}_0,\ldots,\mathbf{f}_m\}$. 
To propagate global semantics from the coarse level $\mathbf{x_m}$ back to the original resolution $\mathbf{x_0}$, we adopt a spatial propagation operation $\mathcal{F}$, which spreads information according to spatial proximity:
\begin{equation} 
    \mathbf{\hat{f}}_{i-1} = \mathcal{F}(\mathbf{x}_{i-1},\mathbf{x}_{i},\mathbf{\hat{f}}_{i}),
    \label{spatial}
\end{equation}

This yields propagated features $\{\mathbf{\hat{f}}_0,\ldots,\mathbf{\hat{f}}_m\}$, which encodes multi-granular geometric information. Intuitively, $\mathbf{\hat{f}}_m$ captures global structural context with the largest receptive field, while $\mathbf{\hat{f}}_0$ preserves detailed local geometry. These multi-granular features are later fused with prompt tokens in the Prompt Propagation module.

To introduce fine-grained geometric adjustment, we feed the point-level features $\mathbf{\hat{f}}_0$ into the shift head and predict coordinate shifts $ {\Delta}\mathbf{x} \in \mathbb{R}^{N\times3}$. The shifted point cloud is accessed by: 
\begin{equation} 
    \mathbf{x_s} = \mathbf{x}_0 + {\alpha_s}{\cdot}{\Delta}\mathbf{x},
\end{equation}
where ${\alpha_s}$ is a scaling factor controlling the magnitude of the point shift. While these point-wise shifts enable continuous, fine-grained deformation of the input geometry, they primarily operate locally.
To further enhance the model’s ability to perceive and emphasize structurally critical regions, we additionally introduce explicit instance-specific keypoints as discrete point-level prompts.

% \noindent\textbf{Feature Propagation.}
\textit{Spatial Propagation:}
The spatial propagation operator $\mathcal{F}$ in Equation~\ref{spatial} interpolates features from center points to neighboring points based on spatial geometry.
Given center coordinates $\{\mathbf{c}_i\}_{i=1}^C \in \mathbb{R}^{C{\times}3}$ with features $\mathbf{f}(\mathbf{c}_i)$ and a neighboring point $\mathbf{n} \in \mathbb{R}^3$, our goal is to estimate $\mathbf{f}(\mathbf{n})$.

We first compute the Euclidean distance:
\begin{equation}
\mathbf{d}(\mathbf{n},\mathbf{c}_i)=||\mathbf{n}-\mathbf{c}_i||.
\end{equation}
Then an inverse-distance weight is assigned:
\begin{equation}
\mathbf{w}(\mathbf{n},\mathbf{c}_i)=\frac{1}{\mathbf{d}(\mathbf{n},\mathbf{c}_i)^p},
\end{equation}
where $p=2$ by default. This results in a set of weights ${\mathbf{w}(\mathbf{n},\mathbf{c}_i)}$ for $i = 1,\ldots,C$. We select the top-K weights:
\begin{equation}
{\mathbf{w}(\mathbf{n},\mathbf{c}_j)} = \texttt{Top-K}({\mathbf{w}(\mathbf{n},\mathbf{c}_i)}),
\end{equation}
where $j=1,\ldots,K$, and $K$ is typically set to 6.
The feature of $\mathbf{n}$ is then interpolated as:
\begin{equation}
\mathbf{f}(\mathbf{n})=\frac{\sum_{j=1}^{K} \mathbf{w}(\mathbf{n},\mathbf{c}_j){\cdot}\mathbf{f}(\mathbf{c}_j)}{\sum_{j=1}^{K} \mathbf{w}(\mathbf{n},\mathbf{c}_j)}.
\end{equation}
This process is applied to every neighboring point $\mathbf{n}$, enabling effective propagation of geometric semantics across tokens.

\subsection{Keypoint Prompter}
To introduce explicit geometric cues and assist the pre-trained model capture subtle structural details, we design a Keypoint Prompter that generates instance-specific keypoints as point-level prompts. Unlike abstract prompt tokens, these keypoint prompts retain direct geometric correspondence with the underlying point cloud, guiding the model toward salient local structures and fine-grained geometric variations in inherently irregular and unstructured point cloud data.

Given the input point cloud $\mathbf{x}_0 \in \mathbb{R}^{N\times3}$, our objective is to identify salient local regions and generate auxiliary keypoint prompts that highlight their geometric importance. We first partition the point cloud into candidate regions using farthest point sampling (FPS) and construct their local neighborhoods via k-nearest neighbors (KNN). To assess the semantic importance of each region, we exploit the global geometric feature $\mathbf{\hat{f}}_m$ extracted from the Point Shift Prompter, which provides high-level contextual priors.

Let $\mathbf{c_s} \in \mathbb{R}^{S \times 3}$ denote the center coordinates of the sampled candidate regions. We concatenate $\mathbf{c_s}$ with the global feature $\mathbf{\hat{f}}_m$ and feed them into an MLP $\mathbf{\Phi}_k$ to simultaneously predict candidate keypoints and their saliency scores:
\begin{equation} 
     \mathbf{{p}_s}, \mathbf{{s}} = \mathbf{\Phi}_k(\mathbf{{c}_s}, \mathbf{\hat{f}}_m),
\end{equation}
where $\mathbf{{p}_s} \in \mathbb{R}^{S\times3}$ denotes the candidate keypoint coordinates and $\mathbf{{s}} \in \mathbb{R}^{S}$ denotes the corresponding saliency scores. 

Obtained the saliency scores $\mathbf{{s}}$, we then select the top $K$ keypoints as final keypoint prompts according to the prominence of saliency. The process can be formulated as:
\begin{equation} 
     \mathbf{{p}_k} = \texttt{Top-K}(\mathbf{{p}_s}, \mathbf{{s}}),
\end{equation}
where $\mathbf{{p}_k} \in \mathbb{R}^{K\times3}$ represents the final set of keypoint prompts, and $K$ is a hyperparameter controlling the number of selected keypoints. These auxiliary keypoints highlight the most informative geometric regions, guiding the model to focus on critical local structures and thus improving adaptation to downstream tasks.

\subsection{Prompt Propagation}
Beyond the fine-grained point-level prompts, we introduce a Prompt Propagation mechanism to inject multi-granular geometric cues throughout the feature extraction hierarchy. 
Given the shifted point cloud $\mathbf{x_s} \in \mathbb{R}^{N\times3}$ and the keypoint prompts $\mathbf{p_k} \in \mathbb{R}^{K\times3}$, we concatenate them and embed the resulting points into patch tokens $\mathbf{h}_0 \in \mathbb{R}^{L{\times}D}$, which are subsequently processed by the frozen pre-trained Transformer blocks following Equations~\ref{attn} and \ref{ffn}.

To guide the model during adaptation, we prepend $P$ learnable prompt tokens $\mathbf{t}_i \in \mathbb{R}^{P{\times}D}$ to the patch tokens at each block. However, these randomly initialized tokens have no inherent geometric meaning, making them difficult to optimize purely under downstream supervision.

To address this, we design a prompt fusion strategy that injects explicit geometric priors into the prompt tokens. Specifically, we select $P$ geometric features from the multi-granular features set ${\mathbf{\hat{f}}_0,\ldots,\mathbf{\hat{f}}_m}$ according to a granularity schedule $\mathcal{G}$, concatenate them into $\mathbf{\hat{f}_g}$, and fuse them with the learnable tokens:
\begin{equation}
\mathbf{\hat{t}}_i = \mathbf{t}_i + \beta_f \cdot \mathbf{\hat{f}_g},
\end{equation}
where $\beta_f$ controls the blending ratio. The fused prompts $\mathbf{\hat{t}}_i\in \mathbb{R}^{P{\times}D}$ then interact with patch tokens via cross-attention, enriching the model with geometry-aware priors at multiple receptive fields.

To further strengthen geometric information flow across the token hierarchy, the Prompt Propagation mechanism incorporates spatial neighborhood relationships. For patch coordinates $\mathbf{x_h} \in \mathbb{R}^{L\times 3}$, we obtain $C$ coarse centers $\mathbf{x_c}$ via FPS and index their corresponding token features $\mathbf{h}_i^c$.

Next, we inject the fused prompt tokens into the patch tokens using a random replacement operation $\mathcal{I}$:
\begin{equation}
\mathbf{h}_i^{\prime} = \mathcal{I}(\mathbf{h}_i, \mathbf{\hat{t}}_i),
\label{inject}
\end{equation}
which functions similarly to dropout~\cite{dropout}, improving robustness during adaptation. The updated tokens are then refined through a spatial propagation operation:
\begin{equation}
\mathbf{h}_i^{\prime\prime} = \mathcal{F}(\mathbf{x_h}, \mathbf{x_c}, \mathbf{h}_i^{\prime}),
\label{propagate}
\end{equation}
yielding geometry-integrated features ${\mathbf{h}_i^{\prime\prime}} \in \mathbb{R}^{L\times D}$.

To mitigate over-smoothing, a lightweight non-linear mapping is applied:
\begin{equation}
\mathbf{\hat{h}}_{i} = \sigma(\mathbf{h}_i^{\prime\prime}{\cdot}\mathbf{W}_{1}){\cdot}\mathbf{W}_{2}+\mathbf{h}_i^{\prime\prime},
\end{equation}
where $\mathbf{W}_{1} \in \mathbb{R}^{d{\times}r}$, $\mathbf{W}_{2} \in  \mathbb{R}^{r{\times}d}$, and $r$ controls the bottleneck dimension. The resulting tokens $\mathbf{\hat{h}}{i}$ are forwarded to subsequent blocks.

Crucially, the propagation module is most effective when paired with geometry-aware prompt tokens. Geometric enhancement allows center-feature propagation to carry meaningful structural information through the token hierarchy, substantially boosting adaptation performance.

% \noindent\textbf{Prompt Injection.}
\textit{Prompt Injection:}
We detail the prompt injection $\mathcal{I}$ mechanism corresponding to Equation~\ref{inject}. Given the input tokens and prompt tokens $[\mathbf{h}_i; \mathbf{\hat{t}}_i] \in \mathbb{R}^{(L+P)\times D}$ in the $i$-th block, we extract the local geometric structure via FPS and KNN. The center tokens $\mathbf{h}_i^c \in \mathbb{R}^{C\times D}$ are indexed according to the sampled center positions $\mathbf{x_c}$, where $C$ denotes the number of centers.

Inspired by dropout~\cite{dropout}, prompt injection randomly substitutes parts of the center and neighboring tokens with geometry-enhanced prompt tokens, facilitating global feature propagation. Owing to the stochasticity of FPS, the permutation of sampled centers varies even when the sets remain consistent. Leveraging this randomness, we adopt a replacement strategy: the last $P$ elements of both $\mathbf{h}_i^c$ and $\mathbf{h}_i$ are replaced with prompt tokens $\mathbf{\hat{t}}_i$, directly mixing prompt information into local regions.

Alternatively, we propose a permutation strategy. We prepend $\mathbf{\hat{t}}_i$ to the full token set $\mathbf{h}_i$ and remove its last $P$ tokens. The fused sequence is then indexed by $\mathbf{x_c}$ and $\mathbf{x_h}$, producing center tokens $\mathbf{h}_i^c$ that naturally incorporate prompt tokens. This approach avoids explicit replacement and injects prompts implicitly through the indexing operation.

\section{{Analysis and Discussion}}
\label{sec-analysis-discussion}
{This section connects the attention-level decomposition of prompt integration with the optimal transport perspective developed earlier, and offers an optimal-transport-inspired analysis and discussion for understanding the behavior of multi-granular geometry-aware prompting.}

\subsection{Attention-level Decomposition of Prompt Integration}
We begin with the standard attention output at input token index $i$:
\begin{equation}
\mathbf{o}_{i} = \texttt{Attn}(\mathbf{W}_Q\mathbf{h}_i,\;\mathbf{W}_K\mathbf{h}_i,\;\mathbf{W}_V\mathbf{h}_i),
\label{eq:attn_no_prompt}
\end{equation}
and the corresponding output when prompt tokens $\hat{\mathbf{t}}_{i}=\{\hat{\mathbf{t}}_{ip}\}_{p=1}^P$ are injected into the key and value spaces:
\begin{equation}
\hat{\mathbf{o}}_{i}
=
\texttt{Attn}(\mathbf{W}_Q\mathbf{h}_i,\;
\mathbf{W}_K[\hat{\mathbf{t}}_{i},\mathbf{h}_i],\;
\mathbf{W}_V[\hat{\mathbf{t}}_{i},\mathbf{h}_i]).
\label{eq:attn_with_prompt}
\end{equation}
Following the linearization used in prior prompt analyses~\cite{prompt}, $\hat{\mathbf{o}}_{i}$ becomes a prompt-weighted interpolation of the original output:
\begin{equation}
\hat{\mathbf{o}}_{i}
=
\sum_{p=1}^P A_{ip}\,\mathbf{W}_V\hat{\mathbf{t}}_{ip}
+ \Big(1-\sum_{p=1}^P A_{ip}\Big)\mathbf{o}_i,
\label{eq-attn_interp}
\end{equation}
where $A_{ip}\!\in\![0,1]$ is the normalized attention weight assigned to the $p$-th prompt. Two observations follow:

\subsubsection{Interpolation view}
Soft prompting performs a convex combination between the backbone output $\mathbf{o}_i$ and prompt-induced corrections, producing a localized offset in output space while preserving the backbone mapping structure.

\subsubsection{Offset–subspace adaptation}
{Prompt integration perturbs backbone features through constrained local shifts, which can be viewed as transporting an empirical feature distribution toward a task-aligned one.} 
The term $\sum_{p=1}^P A_{ip}\,\mathbf{W}_V\hat{\mathbf{t}}_{ip}$ spans a low-dimensional affine subspace parameterized by prompt atoms and attention weights. Prompt learning therefore corresponds to learning localized, low-capacity corrective maps.

\subsection{Connecting Attention Interpolation to Optimal Transport}
We interpret pretrained backbone features
$\{z_i=\mathbf{h}_i\}$
as samples of an empirical source measure
$
\mu=\frac{1}{N}\sum_{i=1}^{N}\delta_{z_i},
$
where $\delta_{z_i}$ denotes a Dirac mass at feature $z_i$.
Likewise, task-aligned features
$\{t_j\}_{j=1}^{M}$
define a target measure
$\nu=\frac{1}{M}\sum_{j=1}^{M}\delta_{t_j}.$
A coupling $\pi\in\Pi(\mu,\nu)$ specifies how mass from $z_i$ is assigned to $t_j$.
Attention weights $A_{ip}$ introduce prompt atoms as intermediate basis elements, yielding a two-stage transport process:
\begin{equation}
\mu
~\xrightarrow{\text{prompt-driven local shift}}~
\tilde{\mu}
~\xrightarrow{\text{residual alignment}}~
\nu.
\end{equation}

We formalize the prompt-integrated output as the following transport-style mapping:
\begin{equation}
T_{\phi}(z_i)=\hat{\mathbf{o}}_i
=
\Big(1-\sum_{p=1}^{P}A_{ip}\Big)\mathbf{o}_i
+\sum_{p=1}^{P}A_{ip}\,\mathbf{W}_{V}\hat{\mathbf{t}}_{ip},
\label{eq-Tphi_definition}
\end{equation}
where $\mathbf{o}_i$ is the backbone output for sample $z_i$, $A_{ip}$ is the normalized attention weight assigned to prompt atom $p$, and $\hat{\mathbf{t}}_{ip}$ denotes the learnable prompt embedding participating in the correction for $z_i$. Under this view, $T_{\phi}$ is exactly the prompt-integrated output in Equation~\ref{eq-attn_interp}, interpreted here as a feature-space transport-style map.
This induces a transformed empirical measure: 
\begin{equation}
\tilde{\mu}=(T_{\phi})_{\ast}\mu,
\end{equation}
where $(T_\phi)_{\ast}$ is the pushforward measure obtained by applying $T_\phi$ to all samples in~$\mu$.
{Here, $\nu$ denotes a conceptual target distribution representing task-aligned features, introduced to facilitate a post-hoc interpretation of adaptation behavior.}
The alignment between $\tilde{\mu}$ and $\nu$ can be expressed using entropic Optimal Transport:
\begin{equation}
W_{\epsilon}(\tilde{\mu},\nu)
=
\min_{\pi\in\Pi(\tilde{\mu},\nu)}
\langle \pi, C\rangle
+
\epsilon\,\mathrm{KL}\big(\pi\,\|\,\tilde{\mu}\otimes\nu\big),
\label{eq-sinkhorn_pushforward}
\end{equation}
where matrix $C\in\mathbb{R}^{N\times M}$ collects pairwise transport costs between transformed source samples and target samples, which has pairwise transport cost $C_{ij}=c(T_{\phi}(z_i),t_j)$. 
The $\Pi(\tilde{\mu},\nu)$ represents the feasible set, $\tilde{\mu}\otimes\nu$ is the independent joint measure of $\tilde{\mu}$ and $\nu$, and $\langle\cdot\rangle$ denotes inner product operation.

This leads to two qualitative implications:
\subsubsection{Localized mass transport}
{
{
Since $T_{\phi}(z_i)$ introduces small attention-weighted corrections, the induced transport can be interpreted as emphasizing manifold-aligned directions encoded by prompt atoms. From this perspective, the resulting coupling may favor semantically consistent correspondences.}

\subsubsection{Low-capacity, structure-preserving map}
{The transport-style map $T_{\phi}$ can be viewed as restricting the corrections to a low-capacity, geometry-aware form, which offers an intuitive way to discuss how pretrained geometric structure may be retained during adaptation.}
}

Hence, attention interpolation in Equation~\ref{eq-attn_interp} can be interpreted as a constrained transport-style correction, where prompt atoms induce intermediate local shifts before alignment toward task-specific features.

% \subsubsection{Localized mass transport}
% {
% {Since $T_{\phi}(z_i)$ introduces small attention-weighted corrections, the induced transport can be interpreted as emphasizing manifold-aligned directions encoded by prompt atoms. From this perspective, the resulting coupling may favor semantically consistent correspondences.}
% \subsubsection{Low-capacity, structure-preserving transport family}
% {The mapping family $\{T_{\phi}\}$ can be viewed as constraining the corrections to be smooth and topology-aware, which offers an intuitive way to discuss how pretrained geometry may be retained during adaptation.}

% {Hence, attention interpolation in Equation~\ref{eq-attn_interp} can be viewed as a learnable approximation to a constrained transport operator. In this interpretation, the attention weights act as prompt-interpolation weights that induce an intermediate local shift through prompt atoms before the transport-style alignment to task-specific features.}
% }

\subsection{Multi-granularity and Hierarchical Transport}

Point-level prompts, including keypoint and point-shift prompts, directly modulate query tokens and instantiate fine-scale corrections in Equation~\ref{eq-Tphi_definition}. The propagation mechanism distributes these corrections across layers, yielding intermediate transformed measures:
\begin{equation}
\mu \xrightarrow{T_{\phi}^{(l)}} \mu^{(l)}
\xrightarrow{T_{\phi}^{(m)}} \mu^{(m)}
\xrightarrow{T_{\phi}^{(g)}} \mu^{(g)},
\end{equation}
%
% where $(l,m,g)$ denote local, meso, and global granularities. This induces an approximate additive OT decomposition:
% \begin{equation}
% W_{c}(\mu,\nu)
% \;\approx\;
% \sum_{k\in\{l,m,g\}}
% W_{c_k}\big((T_{\phi}^{(k)})_{\ast}\mu^{(k-1)},\,\mu^{(k)}\big),
% \label{eq-hier_cost}
% \end{equation}
%

{
where $(l,m,g)$ denote local, meso, and global granularities. This suggests a stage-wise alignment view, where different prompt granularities progressively reduce local, meso-scale, and global mismatches, so that the transformed measure becomes progressively better aligned with the conceptual target distribution $\nu$.}

{
{This stage-wise view is consistent with our empirical findings: the gains from multi-granular prompting, the preferred granularity schedule, and the insertion-layer ablations all support a local-to-global propagation of geometric corrections.}
{It is also supported by the prompt-granularity visualizations in Fig.~\ref{fig-multi-granular-prompts}, where fine-grained prompts focus on localized structures while coarser prompts respond to broader shape layout and global topology, together suggesting that GAPrompt++ benefits from balancing local correction and global context propagation under a low-capacity adaptation scheme.}
}

\section{Experiments Setting}

 We evaluate GAPrompt++ on point cloud classification and segmentation using several pre-trained backbones, including ReCon~\cite{qi2023ReCon}, Point-GPT~\cite{pointgpt}, Point-FEMAE~\cite{zha2023PointFEMAE}, and Point-PQAE~\cite{pointpqae}. In addition, we demonstrate the versatility of our approach by transferring modality-agnostic pre-trained models, including CLIP~\cite{clip} and DINOv3~\cite{dinov3}, to point cloud tasks. {{Besides the standard object-level benchmarks, we additionally include the more challenging large-scale scene semantic segmentation benchmarks~\cite{s3dis,scannet,nuscenes}, as well as outdoor 3D object detection on \textit{NuScenes}~\cite{nuscenes}. These additions extend the empirical study from object-level recognition to large-scale indoor and outdoor scene understanding, enabling a more comprehensive evaluation of scalability and transferability.}} For fair comparison, all baselines adopt the same data augmentation strategy as their corresponding full fine-tuning settings. It should be noted that for datasets or tasks lacking reported results from certain methods, we reproduce their results using the official code to ensure a comprehensive comparison.

\begin{table*}[!ht]
\centering
% \vspace{-5 pt}
\caption{Comparison with traditional supervised learning methods and self-supervised representation learning methods on four point cloud classification datasets. The $^\dagger$ denotes results with a voting strategy~\cite{voting}.}
% \setlength{\extrarowheight}{1pt} % 每行额外增加2pt高度
% \vspace{5 pt}
\begin{small}
\resizebox{\textwidth}{!}{%
\begin{tabular}{l cc ccc ccc}
\toprule
\multirow{2}{*}{\textbf{Method}} & \multirow{2}{*}{\textbf{Publication}} & \multirow{2}{*}{\textbf{Param.(M)} $\downarrow$} & \multicolumn{3}{c}{\textit{\textbf{ScanObjectNN}}}& \multirow{2}{*}{\textit{\textbf{ModelNet40}}} & \multirow{2}{*}{\textit{\textbf{GSModel60}}} & \multirow{2}{*}{\textit{\textbf{uCO3D80}}} \\ \cmidrule{4-6} 
&  &  &  \textit{\textbf{OBJ\_BG}} & \textit{\textbf{OBJ\_ONLY}}  & \textit{\textbf{PB\_T50\_RS}}    &  \\ \midrule \midrule
\multicolumn{9}{c}{\textit{Traditional Supervised Learning Only}}  \\  \midrule
PointNet\cite{pointnet} & \textit{CVPR 17} &3.5& 73.3 & 79.2 & 68.0 & 89.2 &78.86&53.76\\
PointNet++\cite{pointnet++} & \textit{NeurIPS 17} & 1.5 & 82.3 & 84.3 & 77.9 &90.7&86.35&59.60 \\
DGCNN\cite{dgcnn} & \textit{TOG 19} & 1.8 & 82.8 & 86.2 & 78.1 & 92.9 &89.33&56.92\\
% MVTN\cite{mvtn} & \textit{ICCV 21} & 11.2 &  -  &  -  &  -  & 82.8 & 93.8 & - & -\\
PointNeXt\cite{pointnext} & \textit{NeurIPS 22} & 1.4  &87.1&88.6& 87.7 & 94.0 &90.32&65.57\\
PointMLP\cite{pointmlp} & \textit{ICLR 22} & 13.2 &88.5&89.3& 85.4 & 94.5 &90.62&73.63\\
RepSurf\cite{repsurf} & \textit{CVPR 22} & 1.5 &86.1&86.5& 84.3 & 94.4 &89.51&69.32\\
PointVector\cite{pointvector} & \textit{CVPR 23} &1.6&86.8&87.8& 87.8 &93.0&89.77&68.94\\
X-3D\cite{x3d} & \textit{CVPR 24} & 5.4 &87.7&88.4& 90.7 &93.4&89.56&70.65\\
% GPSFormer & \textit{ECCV 24} & 2.4 & - & - & 95.4 & 94.2& - & - \\
MAP\cite{map} & \textit{CVPR 25} & 19.3 &92.8&92.3&92.0& 93.5 &89.29&71.44\\
\midrule 
\multicolumn{9}{c}{\textit{Self-Supervised Representation Learning (Full Fine-Tuning)}}  \\  \midrule
Point-BERT\cite{yu2022PointBERT} & \textit{CVPR 22} & 22.1 & 87.43 & 88.12 & 83.07 & 92.7 &90.39&72.46 \\
% MaskPoint\cite{maskpoint} & \textit{ECCV 22} & 22.1 & 89.70  & 89.30  & 84.60  & 93.8 & - & -  \\  
Point-MAE\cite{pang2022PointMAE} & \textit{ECCV 22} & 22.1 & 90.02 & 88.29 & 85.18 & 93.2 &90.54& 72.73  \\
Point-M2AE\cite{zhang2022PointM2AE} & \textit{NeurIPS 22} & 15.3 & 91.22 & 88.81 & 86.43 & 93.4 &91.05&73.93 \\
ReCon\cite{qi2023ReCon} & \textit{ICML 23} & 43.6 & 94.15 & 93.12 & 89.73 & 93.9 &90.48& 71.26  \\ 
PointGPT-L\cite{pointgpt} & \textit{NeurIPS 23} & 360.5 & 97.20 & 96.60 & 93.40 & 94.1&85.75&65.69\\
Point-FEMAE\cite{zha2023PointFEMAE} & \textit{AAAI 24} & 27.4 & 95.18 & 93.29 & 90.22 & 94.0& 90.46&73.11   \\ 
ReCon++$^\dagger$\cite{recon++} & \textit{ECCV 24} & 657.2 & 98.80 & 97.59 & 95.52 & 94.8&86.72&67.63 \\
Point-PQAE\cite{pointpqae} & \textit{ICCV 25}&22.1&95.01&93.63&89.56&94.0&91.22&74.85 \\
\midrule 
\multicolumn{9}{c}{\textit{Self-Supervised Representation Learning (Parameter-Efficient Fine-Tuning)}}  \\  \midrule
\rowcolor{gray!15}
GAPrompt~\cite{ai2025gaprompt} &  \textit{ICML 25} & 2.0 & 98.97 & 96.73 & 94.31 & 96.2 & 91.20& 75.30\\
% \rowcolor{gray!15}
% GAPrompt$^\dagger$~\cite{ai2025gaprompt} &  \textit{ICML 25} & 2.0 & 98.97 & 96.73 & 94.31 & 96.2 & 91.20& 75.30\\
\rowcolor{gray!15}
\textbf{GAPrompt++} &  \textit{This Paper} & \textbf{1.9} & \textbf{98.97} & \textbf{96.90} & \textbf{95.07} & \textbf{96.4} & \textbf{91.61} & \textbf{76.47} \\
\rowcolor{gray!15}
\textbf{GAPrompt++}$^\dagger$ &  \textit{This Paper} & \textbf{1.9} & \textbf{99.31} & \textbf{97.07} & \textbf{95.14} & \textbf{97.1} & \textbf{92.55} & \textbf{78.06} \\

\bottomrule
\end{tabular}
}
\end{small} 
\label{tab-finetuning}
\vspace{-10 pt}
\end{table*}

\subsection{Datasets \label{datasets}}
\noindent\textbf{\textit{ScanObjectNN}.} 
The \textit{ScanObjectNN}~\cite{scanobjectnn} dataset contains 15K real-world objects from 15 categories across three variants, cropped from the indoor reconstruction dataset ScanNet~\cite{scannet}.
The scans retain challenging artifacts such as background clutter, occlusion, and irregular geometry.

\noindent\textbf{\textit{ModelNet40}.}
The \textit{ModelNet40}~\cite{modelnet40} dataset consists of 12K clean CAD models across 40 categories. The point clouds are complete, uniformly sampled, and free of sensor noise, representing an idealized acquisition setting.

\noindent\textbf{\textit{ShapeNetPart}.}
The \textit{ShapeNetPart} dataset is a widely used benchmark for object part segmentation, comprising 15K point-level annotated objects across 16 object categories and 50 part categories.

\begin{table*}[!ht]
\centering
% \vspace{-5 pt}
\caption{Comparison with 3D-specific parameter-efficient fine-tuning methods on point cloud classification tasks across diverse backbones. Overall accuracy (\%) is reported without voting. The \textcolor{blue}{blue subscripts} denote changes compared to full fine-tuning.}
\setlength{\extrarowheight}{-2pt}
% \vspace{5 pt}
\begin{small}
\resizebox{\textwidth}{!}{
\begin{tabular}{ll ccc ccc ccc}
\toprule
&\multirow{2}{*}{\textbf{Method}} & \multirow{2}{*}{\textbf{Publication}} & \multirow{2}{*}{\textbf{Param.(M)} $\downarrow$} & \multirow{2}{*}{\textbf{FLOPs(G)} $\downarrow$} & \multicolumn{3}{c}{\textbf{\textit{ScanObjectNN}}}& \multirow{2}{*}{\textbf{\textit{ModelNet40}} } & \multirow{2}{*}{\textbf{\textit{GSModel60}}} & \multirow{2}{*}{\textbf{\textit{~uCO3D80~}}}   \\ \cmidrule{6-8} %\cmidrule{9-9}
& &  &  &  & \textit{\textbf{~OBJ\_BG~}} & \textit{\textbf{OBJ\_ONLY}}  & \textit{\textbf{PB\_T50\_RS}}    &   & & \\ 
\midrule \midrule        

 &\multicolumn{1}{|l}{Full Fine-Tuning} & - &360.5 & 67.7 & 97.20 & 96.60 & 93.40 & 94.1&85.75&65.69\\
\cmidrule{2-11}
\multirow{15}{*}{\rotatebox{90}{PointGPT-L\cite{pointgpt}}}&\multicolumn{10}{|c}{\textit{Reparameterization-based Parameter-Efficient Fine-Tuning}}  \\  
\cmidrule{2-11}
 &\multicolumn{1}{|l}{PointLoRA\cite{wang2025pointlora}}&\textit{CVPR 25} &2.8&114.5&98.28&96.04&93.11&94.3&84.14&64.32\\
 &\multicolumn{1}{|l}{MoST\cite{han2025most}} & \textit{CVPR 25} &8.0&76.2&97.93&98.28&97.50&96.2&83.88&65.04\\ 
\cmidrule{2-11}
&\multicolumn{10}{|c}{\textit{Adapter-based Parameter-Efficient Fine-Tuning}}  \\  
\cmidrule{2-11}
 &\multicolumn{1}{|l}{DAPT\cite{zhou2024DAPT}} & \textit{CVPR 24} & 4.2 & 71.6 & 98.11 & 96.21&93.02&94.2&85.14&66.94\\           
 &\multicolumn{1}{|l}{PointGST\cite{liang2025pointgst}} & \textit{TPAMI 25} &2.4&68.0&98.97&97.59&94.83&94.8&85.45&67.76\\
 % &\multicolumn{1}{|l}{PMA\cite{zha2025pma}} & \textit{CVPR 25} &4.9&  -  &98.97&96.73&95.18 &94.9& - & -\\
\cmidrule{2-11}
&\multicolumn{10}{|c}{\textit{Prompt-based Parameter-Efficient Fine-Tuning}}  \\  
\cmidrule{2-11}
&\multicolumn{1}{|l}{IDPT\cite{idpt}} & \textit{ICCV 23} & 10.0 & 75.2 & 98.11 & 96.04 & 92.99 & 93.4 &84.83&66.30\\ 
&\multicolumn{1}{|l}{Point-PEFT\cite{tang2024Point-PEFT}}&\textit{AAAI 24}&3.1&73.2&97.76&96.21&93.11&93.9&85.09&67.54\\  
&\multicolumn{1}{|l}{PPT\cite{zhang2024ppt}} &\textit{MM 25}&5.6&140.7&97.76&96.04&94.10&93.6&84.78&66.88\\
\rowcolor{gray!15}
\cellcolor{white}
&\multicolumn{1}{|l}{GAPrompt\cite{ai2025gaprompt}} &  \textit{ICML 25}&2.0&71.8&98.97&96.73&94.31 &96.2&85.45&67.31\\
\rowcolor{gray!15}
\cellcolor{white}
&\multicolumn{1}{|l}{\textbf{GAPrompt++}} & \textit{This Paper} & \textbf{1.9}\rlap{\textcolor{blue}{$_{99\%\downarrow}$}}
&\textbf{71.6}\rlap{\textcolor{blue}{$_{5.8\%\uparrow}$}}
& \textbf{98.97}\rlap{\textcolor{blue}{$_{1.77\uparrow}$}}
& \textbf{96.90}\rlap{\textcolor{blue}{$_{0.30\uparrow}$}}
& \textbf{95.07}\rlap{\textcolor{blue}{$_{1.67\uparrow}$}}
& \textbf{96.4}\rlap{\textcolor{blue}{$_{2.3\uparrow}$}}
& \textbf{86.94}\rlap{\textcolor{blue}{$_{1.19\uparrow}$}}
& \textbf{69.78}\rlap{\textcolor{blue}{$_{4.09\uparrow}$}} \\
% &\multicolumn{1}{|l}{\textbf{GAPrompt++}} & \textit{This Paper} 
% &\textbf{1.9}
% &\textbf{71.6}
% &\textbf{98.97}
% &\textbf{96.90}
% &\textbf{95.07}
% &\textbf{96.4}
% &\textbf{86.94}
% &\textbf{69.78} \\
% &\multicolumn{1}{|l}{\textcolor{blue}{$\Delta$ over f.t.}} & \textcolor{blue}{-}
% &\textcolor{blue}{-99.5\%}
% &\textcolor{blue}{+5.8\%}
% &\textcolor{blue}{+1.77}
% &\textcolor{blue}{+0.30}
% &\textcolor{blue}{+1.67}
% &\textcolor{blue}{+2.3}
% &\textcolor{blue}{+1.19}
% &\textcolor{blue}{+4.09} \\
\midrule

 &\multicolumn{1}{|l}{Full Fine-Tuning} & - &27.4&4.8&95.18&93.29&90.22&94.0&90.46&73.11 \\
\cmidrule{2-11}
\multirow{15}{*}{\rotatebox{90}{Point-FEMAE\cite{zha2023PointFEMAE}}}&\multicolumn{10}{|c}{\textit{Reparameterization-based Parameter-Efficient Fine-Tuning}}  \\  
\cmidrule{2-11}
 &\multicolumn{1}{|l}{PointLoRA\cite{wang2025pointlora}}&\textit{CVPR 25}&0.8&8.7&94.49&92.60&88.97 &93.4&90.34&73.67\\
 &\multicolumn{1}{|l}{MoST\cite{han2025most}} & \textit{CVPR 25} &2.3&6.0&94.84&93.12&89.69&93.7&90.64&73.71 \\ 
\cmidrule{2-11}
&\multicolumn{10}{|c}{\textit{Adapter-based Parameter-Efficient Fine-Tuning}}  \\  
\cmidrule{2-11}
 &\multicolumn{1}{|l}{DAPT\cite{zhou2024DAPT}} & \textit{CVPR 24} & 1.1 & 5.0 & 93.98 & 92.25 & 88.51&93.2 &90.77&71.56 \\           
 &\multicolumn{1}{|l}{PointGST\cite{liang2025pointgst}} & \textit{TPAMI 25} &0.6&4.8&94.66&92.94&90.22&93.8&90.95&74.96\\
 % &\multicolumn{1}{|l}{PMA\cite{zha2025pma}} & \textit{CVPR 25} &1.1&  -  &  -  &  -  &  -  &  -  & - & - \\
\cmidrule{2-11}
&\multicolumn{10}{|c}{\textit{Prompt-based Parameter-Efficient Fine-Tuning}}  \\  
\cmidrule{2-11}
&\multicolumn{1}{|l}{IDPT\cite{idpt}} & \textit{ICCV 23}&1.7&7.2&92.94&90.88&88.38&93.4&88.95&73.50\\ 
&\multicolumn{1}{|l}{Point-PEFT\cite{tang2024Point-PEFT}}&\textit{AAAI 24}&0.7&7.0&94.32&92.94&89.35&94.3&90.13&74.24\\  
&\multicolumn{1}{|l}{PPT\cite{zhang2024ppt}}&\textit{MM 25}&1.1&10.8&94.49&93.12&90.08&93.7&89.24&74.01\\
\rowcolor{gray!15}
 \cellcolor{white}
&\multicolumn{1}{|l}{GAPrompt\cite{ai2025gaprompt}}&\textit{ICML 25}&0.6&5.0&95.53&93.63&90.67&94.5&90.70&75.30\\
\rowcolor{gray!15}
\cellcolor{white}
&\multicolumn{1}{|l}{\textbf{GAPrompt++}} & \textit{This Paper} 
&\textbf{0.6}\rlap{\textcolor{blue}{$_{98\%\downarrow}$}}
&\textbf{5.0}\rlap{\textcolor{blue}{$_{4.2\%\uparrow}$}}
& \textbf{95.70}\rlap{\textcolor{blue}{$_{0.52\uparrow}$}}
& \textbf{94.49}\rlap{\textcolor{blue}{$_{1.20\uparrow}$}}
& \textbf{90.77}\rlap{\textcolor{blue}{$_{0.55\uparrow}$}}
& \textbf{94.6}\rlap{\textcolor{blue}{$_{0.6\uparrow}$}}
& \textbf{91.28}\rlap{\textcolor{blue}{$_{0.82\uparrow}$}}
& \textbf{76.47}\rlap{\textcolor{blue}{$_{3.36\uparrow}$}} \\
\midrule

&\multicolumn{1}{|l}{Full Fine-Tuning} & - &22.1&4.8&95.01&93.63&89.56&94.0&91.22&73.54\\
\cmidrule{2-11}
\multirow{14}{*}{\rotatebox{90}{Point-PQAE\cite{pointpqae}}}&\multicolumn{10}{|c}{\textit{Reparameterization-based Parameter-Efficient Fine-Tuning}}  \\  
\cmidrule{2-11}
 &\multicolumn{1}{|l}{PointLoRA\cite{wang2025pointlora}}&\textit{CVPR 25}&0.8&8.7&93.98&93.12&88.93&93.4&90.89&72.48\\
 &\multicolumn{1}{|l}{MoST\cite{han2025most}} & \textit{CVPR 25} &2.3&6.0&94.49&93.29&89.35&94.1&91.02&73.96\\ 
\cmidrule{2-11}
&\multicolumn{10}{|c}{\textit{Adapter-based Parameter-Efficient Fine-Tuning}}  \\  
\cmidrule{2-11}
 &\multicolumn{1}{|l}{DAPT\cite{zhou2024DAPT}} & \textit{CVPR 24} & 1.1 & 5.0 &93.46&92.25&87.43&93.1&90.34&73.74\\
 &\multicolumn{1}{|l}{PointGST\cite{liang2025pointgst}} & \textit{TPAMI 25} &0.6&4.8&95.18&93.12&89.14&94.2&90.71&74.00\\
 % &\multicolumn{1}{|l}{PMA\cite{zha2025pma}} & \textit{CVPR 25} &1.1&  -  &  -  &  -  &  -  &  - & - & - \\
\cmidrule{2-11}
&\multicolumn{10}{|c}{\textit{Prompt-based Parameter-Efficient Fine-Tuning}}  \\  
\cmidrule{2-11}
&\multicolumn{1}{|l}{IDPT\cite{idpt}} & \textit{ICCV 23} &1.7&7.2&93.80&93.29&88.62&93.3&90.49&73.89\\ 
&\multicolumn{1}{|l}{Point-PEFT\cite{tang2024Point-PEFT}}&\textit{AAAI 24}&0.7&7.0&94.49&93.80&89.42&94.1&90.05&73.69\\
&\multicolumn{1}{|l}{PPT\cite{zhang2024ppt}} & \textit{MM 25} &1.1&10.8&94.32&93.46&88.97&93.2&89.85&73.74\\
\rowcolor{gray!15}
\cellcolor{white}
&\multicolumn{1}{|l}{GAPrompt\cite{ai2025gaprompt}} &  \textit{ICML 25}&0.6&5.0&95.01&94.15&89.69&94.3&91.20&73.95\\
\rowcolor{gray!15}
\cellcolor{white}
&\multicolumn{1}{|l}{\textbf{GAPrompt++}} &  \textit{This Paper} 
&\textbf{0.6}\rlap{\textcolor{blue}{$_{97\%\downarrow}$}}
&\textbf{5.0}\rlap{\textcolor{blue}{$_{4.2\%\uparrow}$}}
&\textbf{95.35}\rlap{\textcolor{blue}{$_{0.34\uparrow}$}}
&\textbf{94.66}\rlap{\textcolor{blue}{$_{1.03\uparrow}$}}
&\textbf{90.63}\rlap{\textcolor{blue}{$_{1.07\uparrow}$}}
&\textbf{94.6}\rlap{\textcolor{blue}{$_{0.6\uparrow}$}}
&\textbf{91.61}\rlap{\textcolor{blue}{$_{0.39\uparrow}$}}
& \textbf{74.53}\rlap{\textcolor{blue}{$_{0.99\uparrow}$}} \\
\bottomrule
\end{tabular}
}
\end{small} 
\label{tab-parameter-efficient}
\vspace{-10 pt}
\end{table*}

\noindent\textbf{\textit{GSModel60}.}
We curate the \textit{GSModel60} dataset for point cloud classification by systematically collecting and organizing 14K instances across 60 categories from ShapeSplat~\cite{shapesplat} and MACGS~\cite{macgs}.
Unlike existing CAD or scan-based benchmarks, \textit{GSModel60} consists of point clouds sampled from 3D Gaussian primitives reconstructed via Gaussian Splatting from calibrated multi-view images. 
As illustrated in Fig.~\ref{fig-gsmodel60}, this acquisition paradigm produces point clouds with distinctive characteristics.
The point clouds exhibit uneven spatial density, coarse on planar surfaces and dense along thin structures, and inevitably contain reconstruction artifacts inherent to the rendering-based supervision.
Serving as a challenging and complementary benchmark, \textit{GSModel60} better reflects the geometric properties of modern reconstruction pipelines.

\noindent\textbf{\textit{uCO3D80}.}
We construct the \textit{uCO3D80} dataset by filtering out extremely low-quality samples and long-tail categories from uCO3D~\cite{uco3d}, yielding a reliable benchmark for point cloud classification. The point clouds originate from uncalibrated image collections and are reconstructed via a Structure-from-Motion (SfM) pipeline followed by dense Multi-View Stereo (MVS) fusion. As shown in Fig.~\ref{fig-uco3d80}, these reconstructions exhibit characteristic SfM–MVS artifacts, including accumulated geometric drift, incomplete or noisy surfaces, and scattered background fragments, making them substantially more challenging than CAD or scan-based datasets.
Given the broad deployment of MVS in real-world 3D reconstruction scenarios, evaluation on \textit{uCO3D80} provides a more realistic assessment of model robustness and generalizability.

{
\noindent{\textbf{\textit{S3DIS}.}}
{The \textit{S3DIS}~\cite{s3dis} dataset is a widely used indoor scene semantic segmentation benchmark containing densely annotated room-scale point clouds from large office environments.}

\noindent{\textbf{\textit{ScanNet}.}}
{The \textit{ScanNet}~\cite{scannet} dataset is a real-world indoor scene benchmark with semantic annotations for cluttered household and office environments.}

\noindent{\textbf{\textit{NuScenes}.}}
{The \textit{NuScenes}~\cite{nuscenes} dataset is a large-scale autonomous driving benchmark with outdoor LiDAR point clouds, widely used for scene-level segmentation and 3D object detection in urban traffic scenarios.}
}

\begin{table}[t]
\centering
% \vspace{-5 pt}
\caption{Parameter-efficient adaptation of multi-modal pre-trained models on point cloud classification datasets.}
\setlength{\tabcolsep}{1pt}
% \vspace{5 pt}
% \begin{small}
\resizebox{\linewidth}{!}{%
\begin{tabular}{l ccc ccc}
\toprule
\textbf{Method} & \textbf{Modality} & \textbf{Param.(M)} $\downarrow$ & \textit{\textbf{~ModelNet40~}} & \textit{\textbf{~~GSModel60~~}} & \textit{\textbf{~~~uCO3D80~~~}}
\\ \midrule \midrule
Point-BERT\cite{yu2022PointBERT} & 3D & 22.1 & 92.7 &90.39 &72.46 \\  
Point-MAE\cite{pang2022PointMAE} & 3D & 22.1 & 93.2 &90.54 &72.73\\  
Point-M2AE\cite{zhang2022PointM2AE} & 3D &15.3 & 93.4 &91.05 &73.93 \\
PointCLIP\cite{pointclip} & Text+2D & 155.1 &90.0&86.42 & 69.20  \\
CG3D\cite{cg3d} & Text+2D & 164.0 &93.4& 87.75 & 69.89  \\
ACT\cite{ACT_rotate} & 2D+3D & 22.1 &93.5& 90.88 & 72.47  \\
ReCon\cite{qi2023ReCon} & Text+2D+3D & 43.6 & 93.9 &90.48 & 71.26\\ 

\midrule 
DINOv3 & 2D & 86.3 &92.0 &86.49&68.42 \\
\midrule
w.IDPT\cite{idpt} & 2D & 4.6 &89.5&84.87&66.33 \\ 
w.Point-PEFT\cite{tang2024Point-PEFT} & 2D & 2.6 & 90.9&86.21&67.82\\
w.PPT\cite{zhang2024ppt} & 2D & 3.4 &90.2&85.49&66.74\\
\rowcolor{gray!15}
w.GAPrompt\cite{ai2025gaprompt} & 2D & 2.0 &91.7&87.35&68.96\\
\rowcolor{gray!15}
w.\textbf{GAPrompt++} 
& 2D
& \textbf{1.9}\rlap{\textcolor{blue}{$_{98\%\downarrow}$}}
&\textbf{93.8}\rlap{\textcolor{blue}{$_{1.8\uparrow}$}}
&\textbf{89.56}\rlap{\textcolor{blue}{$_{3.07\uparrow}$}}
& \textbf{71.95}\rlap{\textcolor{blue}{$_{3.53\uparrow}$}} \\

\midrule 
CLIP-Lang\cite{clip} & Text &85.9&92.2&87.83&68.50 \\
\midrule
w.IDPT\cite{idpt}  & Text & 4.6 &90.8&85.60&63.34\\ 
w.Point-PEFT\cite{tang2024Point-PEFT} & Text & 2.6 &92.5& 88.51 & 68.15 \\
w.PPT\cite{zhang2024ppt} & Text & 3.4 &91.3& 85.76& 65.82 \\
\rowcolor{gray!15}
w.GAPrompt\cite{ai2025gaprompt} & Text & 2.0 &93.2&89.96&71.24\\
\rowcolor{gray!15}
w.\textbf{GAPrompt++} & Text & \textbf{1.9}\rlap{\textcolor{blue}{$_{98\%\downarrow}$}}
& \textbf{96.2}\rlap{\textcolor{blue}{$_{4.0\uparrow}$}}
&\textbf{91.63}\rlap{\textcolor{blue}{$_{3.80\uparrow}$}}
&\textbf{77.54}\rlap{\textcolor{blue}{$_{9.04\uparrow}$}}\\

\bottomrule
\end{tabular}
}
% \end{small}
\label{multi-modal-adaptation}
\end{table}

\section{Quantitative Analysis}
\subsection{Comparison with Supervised and Self-supervised Methods}
Table~\ref{tab-finetuning} presents a comprehensive comparison against state-of-the-art supervised and self-supervised approaches across four classification benchmarks, including three variants of \textit{ScanObjectNN}.

Overall, self-supervised pre-training consistently outperforms purely supervised learning, as it provides more transferable representations. However, these gains are often achieved at significantly higher adaptation cost. For instance, the trainable parameters increase dramatically from 22.1M in Point-BERT~\cite{yu2022PointBERT} to 657.2M in ReCon++~\cite{recon++}, whereas lightweight supervised baselines such as PointNeXt~\cite{pointnext} require only 1.4M parameters.

Our method achieves state-of-the-art performance with extremely compact adaptation.
GAPrompt attains nearly saturated accuracy, reaching \textbf{98.97\%} on \textit{OBJ\_BG} and \textbf{96.2\%} on \textit{ModelNet40}.
The enhanced GAPrompt++ further improves the results to \textbf{99.31\%} on \textit{OBJ\_BG}, \textbf{97.1\%} on \textit{ModelNet40}, \textbf{92.55\%} on \textit{GSModel60}, and \textbf{78.06\%} on \textit{uCO3D80}, while requiring only \textbf{1.9M} trainable parameters for backbone tuning. It is worth noting that the best performance on \textit{ScanObjectNN} and \textit{ModelNet40} is achieved when equipped with PointGPT as backbone, whereas \textit{GSModel60} benefits from PointPQAE and \textit{uCO3D80} leverages Point-FEMAE.
These results collectively indicate that multi-granular, geometry-aware prompting delivers superior accuracy together with high parameter efficiency. Its capability to jointly capture fine-scale structural cues and global semantic patterns enables effective task alignment with a compact adaptation budget.

It is worth noting that commonly used benchmarks such as \textit{ScanObjectNN} and \textit{ModelNet40} have become saturated, with recent models like Point-GPT~\cite{pointgpt} and ReCon++~\cite{recon++} reaching over 98\% and 96\% accuracy respectively. Their impressive performance largely stems from massive model capacity and access to large-scale curated post-training datasets. However, due to limited category diversity and idealized acquisition conditions, these benchmarks provide diminishing discriminatory power for evaluating future progress. In comparison, our curated \textit{GSModel60} and \textit{uCO3D80} benchmarks uncover substantial performance gaps that remain unresolved in real-world reconstruction contexts. Models like ReCon~\cite{qi2023ReCon} and Point-PQAE~\cite{pointpqae} exhibit stronger generalization than Point-GPT~\cite{pointgpt}, since point clouds derived from 3D Gaussian splatting and multi-view reconstruction deviate considerably from the annotated CAD and RGB-D distributions on which Point-GPT is post-trained. These observations underscore the necessity of challenging, reconstruction-oriented evaluation settings and corroborate the practical relevance of our proposed datasets.

\begin{table}[t]
\centering
\caption{Comparison with recent parameter-efficient fine-tuning methods on object part segmentation benchmarks. $^{*}$ denotes results reproduced using the official implementation.}
\setlength{\extrarowheight}{-1pt}
\begin{small}
\resizebox{\linewidth}{!}{
\begin{tabular}{llcccc}
\toprule
&\textbf{Methods} & \textbf{Publication} & \textbf{Param.(M)} & \textbf{mIoU$_C$(\%)} & \textbf{mIoU$_I$(\%)} \\
\midrule
\midrule
&\multicolumn{1}{|l}{Full Fine-Tuning} & - & 27.06 & 84.52 & 86.1 \\
\cmidrule{2-6}
\multirow{13}{*}{\rotatebox{90}{ReCon\cite{qi2023ReCon}}}&\multicolumn{5}{|c}{\textit{Reparameterization-based Methods}}  \\ 
\cmidrule{2-6}
&\multicolumn{1}{|l}{PointLoRA\cite{wang2025pointlora}} & \textit{CVPR 25} & 5.63 & 83.98 & 85.4 \\
&\multicolumn{1}{|l}{MoST\cite{han2025most}} & \textit{CVPR 25} &5.81&84.42&86.0\\
\cmidrule{2-6}
&\multicolumn{5}{|c}{\textit{Adapter-based Methods}}  \\ 
\cmidrule{2-6}
&\multicolumn{1}{|l}{DAPT\cite{zhou2024DAPT}} & CVPR 24 & 5.65 & 83.87 & 85.7 \\
&\multicolumn{1}{|l}{PointGST\cite{liang2025pointgst}} & \textit{TPAMI 25} & 5.59 & 83.98 & 85.8 \\
% &\multicolumn{1}{|l}{PMA} & \textit{CVPR 25} & 5.64 & 84.00 & 86.1 \\
\cmidrule{2-6}
&\multicolumn{5}{|c}{\textit{Prompt-based Methods}}  \\ 
\cmidrule{2-6}
&\multicolumn{1}{|l}{IDPT\cite{idpt}} & \textit{ICCV 23} & 5.69 & 83.66& 85.7 \\
&\multicolumn{1}{|l}{Point-PEFT\cite{tang2024Point-PEFT}} & \textit{AAAI 24} & 5.62 & 83.10 & 85.1 \\
&\multicolumn{1}{|l}{PPT$^{*}$\cite{zhang2024ppt}} & \textit{MM 25} & 5.62 & 83.45 & 85.5 \\
\rowcolor{gray!15}
\cellcolor{white}
&\multicolumn{1}{|l}{GAPrompt\cite{ai2025gaprompt}} & \textit{ICML 25} & 5.60 & 83.42 & 85.6 \\
\rowcolor{gray!15}
\cellcolor{white}
&\multicolumn{1}{|l}{\textbf{GAPrompt++}} & \textit{This Paper} & \textbf{5.51} & \textbf{{83.61}} & \textbf{85.9} \\

\midrule
&\multicolumn{1}{|l}{Full Fine-Tuning} & - & 27.06 & 84.91 & 86.3 \\
\cmidrule{2-6}
\multirow{13}{*}{\rotatebox{90}{Point-FEMAE\cite{zha2023PointFEMAE}}}&\multicolumn{5}{|c}{\textit{Reparameterization-based Methods}}  \\ 
\cmidrule{2-6}
&\multicolumn{1}{|l}{PointLoRA\cite{wang2025pointlora}} & \textit{CVPR 25} & 5.63 &83.69& 85.6 \\
&\multicolumn{1}{|l}{MoST\cite{han2025most}} & \textit{CVPR 25} &5.81&83.74&85.8\\
\cmidrule{2-6}
&\multicolumn{5}{|c}{\textit{Adapter-based Methods}}  \\ 
\cmidrule{2-6}
&\multicolumn{1}{|l}{DAPT\cite{zhou2024DAPT}} & \textit{CVPR 24} & 5.65 & 83.76 & 85.7 \\
&\multicolumn{1}{|l}{PointGST\cite{liang2025pointgst}} & \textit{TPAMI 25} & 5.59 & 84.04 & 86.0 \\
% &\multicolumn{1}{|l}{PMA} & \textit{CVPR 25} & 5.64 & 84.05 & 86.1 \\
\cmidrule{2-6}
&\multicolumn{5}{|c}{\textit{Prompt-based Methods}}  \\ 
\cmidrule{2-6}
&\multicolumn{1}{|l}{IDPT\cite{idpt}} & \textit{ICCV 23} & 5.69 & 83.75 & 85.8 \\
&\multicolumn{1}{|l}{Point-PEFT\cite{tang2024Point-PEFT}} & \textit{AAAI 24} & 5.62 & 83.25 & 85.5 \\
&\multicolumn{1}{|l}{PPT\cite{zhang2024ppt}} & \textit{MM 25} & 5.62 & 83.64 & 85.6 \\
\rowcolor{gray!15}
\cellcolor{white}
&\multicolumn{1}{|l}{GAPrompt\cite{ai2025gaprompt}} & \textit{ICML 25} & 5.60 &83.84&85.7\\
\rowcolor{gray!15}
\cellcolor{white}
&\multicolumn{1}{|l}{\textbf{GAPrompt++}} & \textit{This Paper} & \textbf{5.51} & \textbf{84.09} & \textbf{86.1} \\

\midrule
&\multicolumn{1}{|l}{Full Fine-Tuning} & - & 27.06 & 84.57 & 86.1 \\
\cmidrule{2-6}
\multirow{13}{*}{\rotatebox{90}{Point-PQAE\cite{pointpqae}}}&\multicolumn{5}{|c}{\textit{Reparameterization-based Methods}}  \\ 
\cmidrule{2-6}
&\multicolumn{1}{|l}{PointLoRA\cite{wang2025pointlora}} & \textit{CVPR 25} & 5.63 &83.34& 85.5 \\
&\multicolumn{1}{|l}{MoST\cite{han2025most}} & \textit{CVPR 25} &5.81&83.50&85.7\\
\cmidrule{2-6}
&\multicolumn{5}{|c}{\textit{Adapter-based Methods}}  \\ 
\cmidrule{2-6}
&\multicolumn{1}{|l}{DAPT\cite{zhou2024DAPT}} & \textit{CVPR 24} & 5.65 & 83.82 & 85.6 \\
&\multicolumn{1}{|l}{PointGST\cite{liang2025pointgst}} & \textit{TPAMI 25} & 5.59 & 83.57 & 85.7 \\
% &\multicolumn{1}{|l}{PMA} & \textit{CVPR 25} & 5.64 & 84.05 & 86.1 \\
\cmidrule{2-6}
&\multicolumn{5}{|c}{\textit{Prompt-based Methods}}  \\ 
\cmidrule{2-6}
&\multicolumn{1}{|l}{IDPT\cite{idpt}} & \textit{ICCV 23} & 5.69 & 83.44 & 85.5 \\
&\multicolumn{1}{|l}{Point-PEFT\cite{tang2024Point-PEFT}} & \textit{AAAI 24} & 5.62 & 83.20 & 85.3 \\
&\multicolumn{1}{|l}{PPT\cite{zhang2024ppt}} & \textit{MM 25} & 5.62 & 83.51 & 85.5 \\
\rowcolor{gray!15}
\cellcolor{white}
&\multicolumn{1}{|l}{GAPrompt\cite{ai2025gaprompt}} & \textit{ICML 25} & 5.60 & 83.45 & 85.5 \\
\rowcolor{gray!15}
\cellcolor{white}
&\multicolumn{1}{|l}{\textbf{GAPrompt++}} & \textit{This Paper} & \textbf{5.51} & \textbf{83.69} & \textbf{85.8} \\
\bottomrule
\end{tabular}
}
\end{small}
\label{tab-part-segmentation}
\end{table}

\subsection{Comparison with 3D Parameter-efficient Methods}
Table~\ref{tab-parameter-efficient} reports comparisons with recent parameter-efficient fine-tuning techniques on four classification benchmarks. Existing approaches can be broadly grouped into reparameterization, adapter-based, and prompting strategies, among which our GAPrompt and GAPrompt++ fall into the latter category. From the perspective of fair comparison, we focus primarily on prompting-based competitors.

Our design introduces multi-granular geometry-aware prompting, which enables consistent advantages over full fine-tuning. This indicates that an appropriately structured parameter-efficient formulation can effectively activate pre-trained representations and sustain a highly discriminative feature space. For instance, GAPrompt++ exceeds full fine-tuning of the recent Point-PQAE~\cite{pointpqae} by \textbf{1.07\%} on \textit{PB\_T50\_RS} and \textbf{0.99\%} on \textit{uCO3D80}, while earlier prompting methods achieve more modest gains on several datasets.

Moreover, GAPrompt++ maintains strong performance across diverse pre-trained backbones, including PointGPT-L~\cite{pointgpt}, Point-FEMAE~\cite{zha2023PointFEMAE}, and Point-PQAE~\cite{pointpqae}. This stability reflects backbone-agnostic behavior under markedly different model sizes, training paradigms, and data sources. On more challenging benchmarks such as \textit{uCO3D80}, GAPrompt++ continues to benefit from jointly modeling local fine-grained geometry and global semantics, which supports stronger adaptation under substantial distribution gaps and reconstruction artifacts.

\subsection{Parameter-efficient Adaptation of Text or 2D Images Models}
Beyond adapting point cloud pre-trained models to downstream tasks, GAPrompt++ further enables the use of pre-trained models from other modalities, removing dependence on costly 3D data collection~\cite{pix4point,any2point}. This demonstrates that our prompting mechanism can unlock cross-modal representational transfer.

As shown in Table~\ref{multi-modal-adaptation}, we evaluate text-only and image-only pre-trained models such as DINOv3~\cite{dinov3} and the language branch of CLIP~\cite{clip}. Their backbone weights are frozen during adaptation, and only the prompting modules, a lightweight point cloud embedder, and a task head are trained. Direct full fine-tuning of these models yields unsatisfactory performance for point cloud classification, likely due to large modality gaps and the lack of 3D priors. However, GAPrompt++ introduces multi-granular geometric awareness through prompting, making the adaptation much more effective.

Remarkably, the language branch of CLIP reaches \textbf{77.54\%} accuracy on the \textit{uCO3D80} dataset using GAPrompt++, outperforming its fully fine-tuned counterpart by \textbf{9.04\%}. This indicates that the geometric cues injected through our prompting design can compensate for missing 3D supervision, and in some cases even rival specialized 3D pre-training. These results highlight the potential of GAPrompt++ as a generic framework for efficient cross-modal transfer into the 3D domain.

\subsection{Experiments on the Part Segmentation Task}
The point cloud segmentation task requires dense point-wise predictions and thus serves as a strong indicator of fine-grained perception capability. As shown in Table~\ref{tab-part-segmentation}, we evaluate our method on the \textit{ShapeNetPart} dataset, which is widely adopted for object part segmentation. Following common practices~\cite{idpt}, we report mean intersection over union across classes (mIoU$_C$) and across instances (mIoU$_I$).

Our experiments consider three recent representative backbones, including ReCon~\cite{qi2023ReCon}, Point-FEMAE~\cite{zha2023PointFEMAE}, and Point-PQAE~\cite{pointpqae}. Although \textit{ShapeNetPart} is close to saturation, with rare methods~\cite{pointnext} exceeding 87\% mIoU$_I$ due to annotation noise and intrinsic ambiguity, our approach still yields a competitive performance of \textbf{86.1\%} mIoU$_I$ on Point-FEMAE. We attribute this to the multi-granular prompting design, which enhances geometric sensitivity and reinforces part-level discrimination. It is also worth noting that the majority of learnable parameters across all compared methods reside in the segmentation head, with our prompt modules contributing only a small additional parameter budget. Even under this setting, GAPrompt++ retains the lowest number of additional trainable parameters while preserving strong segmentation accuracy. This further supports the efficiency and robustness of our prompting framework when adapting to dense prediction tasks.

\begin{table*}[!ht]
\centering
\caption{{Comparison with recent parameter-efficient fine-tuning methods on large-scale scene semantic segmentation benchmarks. Results are reported on \textit{S3DIS} Area-5, \textit{ScanNet} validation set, and \textit{NuScenes} validation set.}}
\label{tab-scene-segmentation}
\vspace{-4pt}
\setlength{\tabcolsep}{6pt}
\begin{small}
\resizebox{\textwidth}{!}{
\begin{tabular}{lcc ccc ccc ccc}
\toprule
\multirow{2}{*}{\textbf{Method}}
&\multirow{2}{*}{\textbf{Publication}}
&\multirow{2}{*}{\textbf{Param.(M)} $\downarrow$}
&\multicolumn{3}{c}{\textbf{\textit{S3DIS Area-5}}}
&\multicolumn{3}{c}{\textbf{\textit{ScanNet Val}}}
&\multicolumn{3}{c}{\textbf{\textit{NuScenes Val}}}
\\
\cmidrule(lr){4-6}
\cmidrule(lr){7-9}
\cmidrule(lr){10-12}
&&&\textbf{mIoU}$\uparrow$&\textbf{mAcc}$\uparrow$&\textbf{allAcc}$\uparrow$&\textbf{mIoU}$\uparrow$&\textbf{mAcc}$\uparrow$&\textbf{allAcc}$\uparrow$&\textbf{mIoU}$\uparrow$&\textbf{mAcc}$\uparrow$&\textbf{allAcc}$\uparrow$\\
\midrule
\midrule
Point Transformer v3~\cite{pointtransformerv3} & \textit{CVPR 24} & 124.8 & 73.4 & 78.9 & 91.7 & 77.6 & 85.0 & 92.0 & 80.4 & 87.2 & 94.7 \\
Sonata (f.t.)~\cite{sonata} & \textit{CVPR 25} & 124.8 & 76.0 & 81.6 & 93.0 & 79.4 & 86.1 & 92.5 & 81.7 & 87.9 & 95.0 \\
Concerto (f.t.)~\cite{concerto} & \textit{NeurIPS 25} & 124.8 & 77.4 & 85.0 & 93.2 & 80.7 & 87.4 & 93.1 & 82.0 & 88.1 & 94.6 \\
Utonia (f.t.)~\cite{utonia} & \textit{ICML 26} & 157.7 & 78.1 & 86.8 & 93.4 & 81.1 & 89.1 & 93.3 & 82.2 & 88.3 & 94.8 \\
\midrule
Concerto (lin.)~\cite{concerto} & \textit{NeurIPS 25} & 0.2 & 73.5 & 81.3 & 90.9 & 77.3 & 86.6 & 91.7 & 74.2 & 83.2 & 93.3 \\

w.PointLoRA~\cite{wang2025pointlora}& \textit{CVPR 25} & 1.9 & 75.2 & 82.4 & 90.1 & 79.5 & 87.0 & 92.2 & 76.1 & 84.2 & 93.3 \\
w.PointGST~\cite{liang2025pointgst}& \textit{TPAMI 25} & 1.1 & 76.1 & 83.1 & 91.9 & 80.2 & 88.0 & 92.8 & 76.3 & 84.6 & 93.5 \\
w.GEM~\cite{gem}& \textit{NeurIPS 25} & 1.8 & 76.2 & 83.2 & 91.6 & 80.3 & 88.1 & 93.0 & 77.6 & 85.7 & 93.7 \\
\rowcolor{gray!15}
w.GAPrompt~\cite{ai2025gaprompt}& \textit{ICML 25} & 1.0 & 75.8 & 82.7 & 91.3 & 79.9 & 87.6 & 92.7 & 77.2 & 85.1 & 93.4 \\
\rowcolor{gray!15}
w.\textbf{GAPrompt++}& \textit{This Paper} & \textbf{1.0} & \textbf{76.5} & \textbf{83.3} & \textbf{92.3} & \textbf{80.5} & \textbf{88.3} & \textbf{93.1} & \textbf{78.6} & \textbf{86.7} & \textbf{94.3} \\

\midrule
Utonia (lin.)~\cite{utonia} &\textit{ICML 26}& 0.2 & 74.7 & 81.2 & 91.5 & 77.7 & 87.0 & 91.9 & 75.5 & 84.8 & 93.7 \\

w.PointLoRA~\cite{wang2025pointlora}& \textit{CVPR 25} & 2.1 & 76.3 & 83.3 & 92.1 & 79.8 & 87.9 & 92.6 & 78.5 & 86.0 & 93.6 \\
w.PointGST~\cite{liang2025pointgst}& \textit{TPAMI 25} & 1.2 & 76.5 & 83.6 & 92.2 & 80.4 & 88.3 & 92.9 & 77.9 & 86.3 & 93.8 \\
w.GEM~\cite{gem}& \textit{NeurIPS 25} & 1.9 & 77.0 & 83.9 & 92.6 & 80.3 & 88.2 & 92.8 & 78.8 & 87.1 & 94.2 \\
\rowcolor{gray!15}
w.GAPrompt~\cite{ai2025gaprompt}& \textit{ICML 25} & 1.0 & 77.2 & 83.9 & 92.7 & 79.9 & 88.0 & 92.7 & 78.4 & 86.6 & 94.1 \\
\rowcolor{gray!15}
w.\textbf{GAPrompt++}& \textit{This Paper} & \textbf{1.0} & \textbf{78.0} & \textbf{84.4} & \textbf{92.8} & \textbf{80.6} & \textbf{88.6} & \textbf{93.1} & \textbf{80.0} & \textbf{87.7} & \textbf{94.7} \\
\bottomrule
\end{tabular}
}
\end{small}
\vspace{-16pt}
\end{table*}

\begin{table}[t]
\centering
% \vspace{-9 pt}
\caption{Comparisons of PEFT methods from NLP and 2D Vision on the hardest variant of ScanObjectNN. 
% All methods choose Point-MAE as the backbone. 
}
% \vspace{5pt}
% \setlength{\extrarowheight}{3pt} % 每行额外增加2pt高度
\resizebox{\linewidth}{!}{
\begin{tabular}{lcccc}
\toprule
\textbf{Method}&\textbf{Domain}&\textbf{Param.(M)$\downarrow$} & \textbf{\textit{ScanObjectNN}}&\textbf{\textit{uCO3D80}}  \\ 
\midrule
\midrule
Point-PQAE\cite{pointpqae} & - & 22.1 &89.56&73.54 \\ 
w.Linear Probing & - & 0.3 & 83.19 & 63.78  \\
\midrule
w.Adapter\cite{adapter} & NLP &0.9& 87.14 & 69.45 \\
w.Prefix Tuning\cite{li2021prefix} & NLP & 0.7 & 85.59 & 67.22 \\ 
w.BitFit\cite{bitfit} & NLP & 0.3 & 84.91 & 67.83 \\
w.LoRA\cite{hu2021lora} & NLP & 0.9 & 86.11 & 68.34 \\
w.DePT\cite{dept} & NLP & 0.3 & 84.05 & 67.42 \\
w.FourierFT\cite{fourierft} & NLP & 0.3 & 83.70 & 66.51 \\
\midrule
w.VPT\cite{vpt} & 2D & 0.4 & 85.93 & 67.25  \\ 
w.AdapterFormer\cite{adaptformer} & 2D & 0.9 & 87.47 & 69.80  \\ 
w.BI-AdaptFormer\cite{biadaptformer} & 2D & 0.4 & 87.65 & 69.72 \\
w.SSF\cite{ssf} & 2D & 0.4 & 86.62 & 68.39  \\ 
w.SCT\cite{sct} & 2D & 0.3 & 85.59 & 67.33  \\
w.InsVP\cite{liu2024insvp} & 2D & 1.3 & 87.47 & 68.97  \\
w.VFPT\cite{vfpt} & 2D & 0.4 & 86.28 & 67.45  \\
\midrule
\rowcolor{gray!15}
w.GAPrompt~\cite{ai2025gaprompt} & 3D & 0.6 & 89.70 & 73.95  \\ 
\rowcolor{gray!15}
w.\textbf{GAPrompt++} & 3D & \textbf{0.6} & \textbf{90.63} & \textbf{74.53} \\ 
\bottomrule
\end{tabular}
}
\label{tab-basic-peft}
\vspace{-10pt}
\end{table}

{
{
\subsection{Experiments on Large-scale Scene Semantic Segmentation}
To further validate the dense prediction capability of GAPrompt++ beyond object-level part segmentation, we evaluate it on large-scale scene semantic segmentation benchmarks, including indoor datasets \textit{S3DIS}~\cite{s3dis} and \textit{ScanNet}~\cite{scannet}, and the outdoor autonomous driving dataset \textit{NuScenes}~\cite{nuscenes}. These benchmarks involve cluttered scenes with over \textbf{100K} points per sample, providing a more challenging setting for assessing geometric understanding, scene-level generalization, and scalability to large point clouds. Our experiments are conducted with two representative self-supervised backbones, Concerto~\cite{concerto} and Utonia~\cite{utonia}.
}

{
As shown in Table~\ref{tab-scene-segmentation}, GAPrompt++ consistently outperforms recent PEFT methods such as PointLoRA~\cite{wang2025pointlora}, PointGST~\cite{liang2025pointgst}, and GEM~\cite{gem} across all three datasets and both scene-level backbones. In terms of mIoU, our method improves over recent PEFT baselines by \textbf{+0.3--1.0\%} on \textit{S3DIS}~\cite{s3dis}, \textbf{+0.2\%} on \textit{ScanNet}~\cite{scannet}, and \textbf{+1.0--1.2\%} on \textit{NuScenes}~\cite{nuscenes}, while using only around 1.0M trainable parameters. Compared with the previous GAPrompt~\cite{ai2025gaprompt}, GAPrompt++ further delivers consistent gains across all three datasets. These results indicate that the proposed multi-granular prompting design effectively captures both local geometric cues and global scene context, further supporting the robustness and scalability of our framework for large-scale dense prediction tasks.
Note that large-scale scene segmentation is substantially more challenging than object-level classification, due to denser point clouds and larger model capacity. Current PEFT methods still cannot fully surpass the adaptation capacity of full fine-tuning, which should be further explored in future work.
}
}

\begin{figure*}[!ht]
\vspace{-8pt}
\centering
\includegraphics[width=0.95\textwidth]{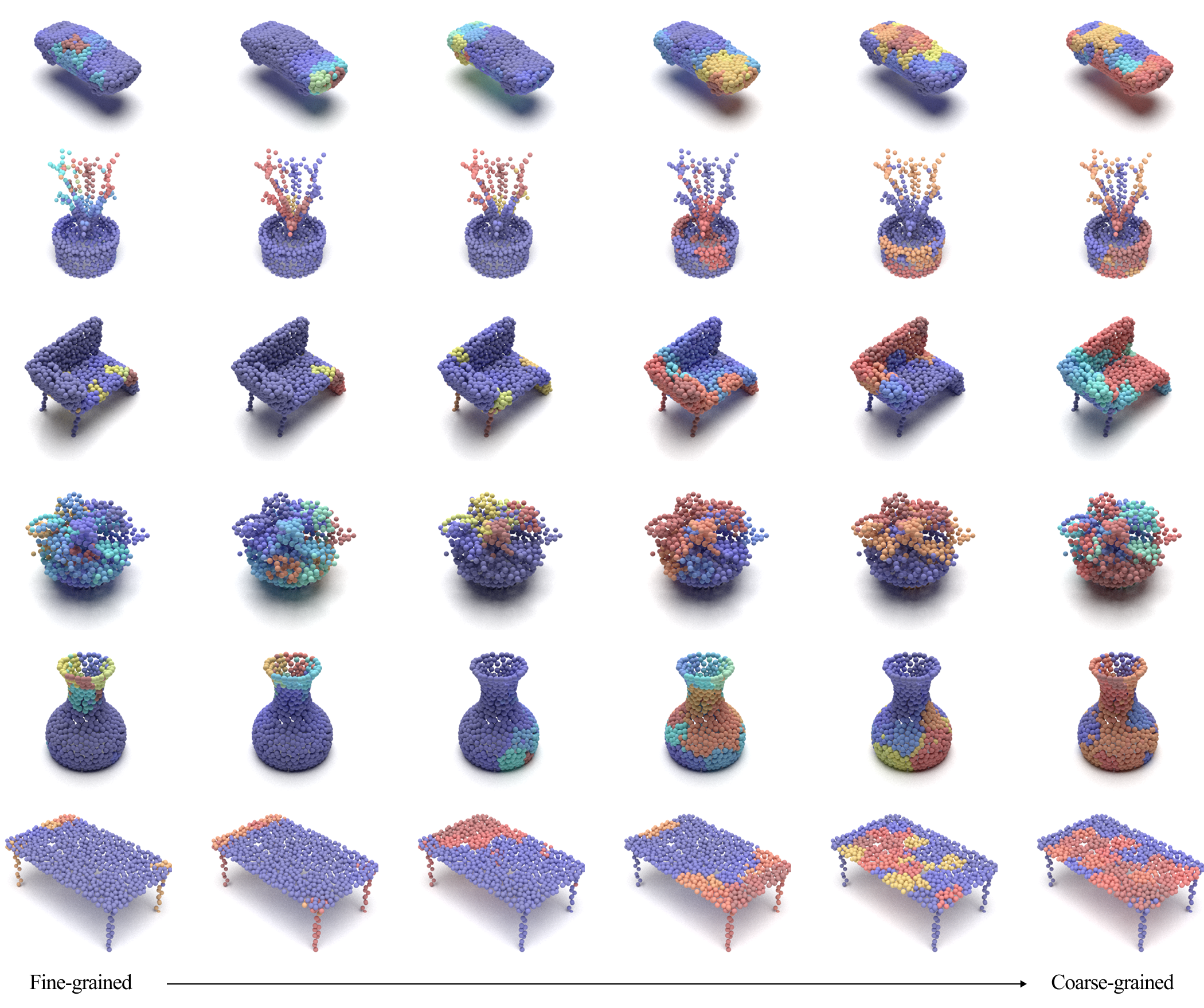}
\vspace{-12pt}
\caption{Visualization of attention distributions induced by our fused multi-granular geometric prompts. Samples are obtained on the \textit{ModelNet40} dataset. Warm colors correspond to higher attention scores.}
\label{fig-multi-granular-prompts}
\vspace{-12pt}
\end{figure*}

\subsection{Comparison with NLP and 2D Parameter-efficient Methods}
To further highlight the necessity of multi-granular geometric cues in prompting, we compare GAPrompt++ against parameter-efficient tuning strategies originally developed for NLP and 2D vision. As shown in Table~\ref{tab-basic-peft}, all methods are evaluated on recent Point-PQAE~\cite{pointpqae} using the \textit{PB\_T50\_RS} split of \textit{ScanObjectNN} and the \textit{uCO3D80} dataset. While these basic PEFT techniques provide moderate gains over linear probing, they still fall noticeably short of full fine-tuning, suggesting that they lack the inductive bias needed to cope with the irregular and anisotropic nature of point clouds.

\begin{table}[!t] 
\centering
% \vspace{-9 pt}
\caption{The effect of components in our GAPrompt++.}
\setlength{\tabcolsep}{1pt}
\begin{scriptsize}
\resizebox{\linewidth}{!}{
\begin{tabular}{c|c|c|cc}
\toprule 
\textbf{Point Shift Prompter}&\textbf{Keypoint Prompter}&\textbf{Prompt Propagation}&\textbf{\textit{ScanObjectNN}}&\textbf{\textit{uCO3D80}} \\ 
\cmidrule(r){1-1} \cmidrule(lr){2-2} \cmidrule(lr){3-3} \cmidrule(l){4-5}
\multicolumn{3}{c|}{Linear Probing of Point-PQAE\cite{pointpqae} (Baseline)}&83.19&63.78 \\
\cmidrule(){1-3} \cmidrule(){4-5}
\checkmark & - & - & 86.55&65.21 \\ 
- & \checkmark & - & 85.83&66.42 \\ 
- & - &\checkmark & 87.96&70.65 \\ 
\checkmark  & \checkmark & - & 88.62&69.85   \\  
\checkmark & \checkmark & \checkmark & \textbf{90.63} & \textbf{74.53}  \\    
\bottomrule
\end{tabular}
}
\end{scriptsize}
\label{tab-ablation-on-main-components}
\vspace{-10pt}
\end{table}

In contrast, GAPrompt++ matches or even exceeds full fine-tuning. By explicitly injecting multi-granular geometric information, it preserves representation discriminability and maintains adaptation stability, leading to consistently stronger performance than transferred NLP or 2D PEFT approaches. Notably, these improvements are achieved with a comparable number of trainable parameters. The compact Point Shift Prompter extracts geometric cues directly from the point cloud and keeps the overall design lightweight. Overall, the results highlight the value of geometry-aware prompting for effective parameter-efficient adaptation in 3D vision.

\begin{figure*}[t]
\vspace{-8pt}
\centering
\includegraphics[width=\textwidth]{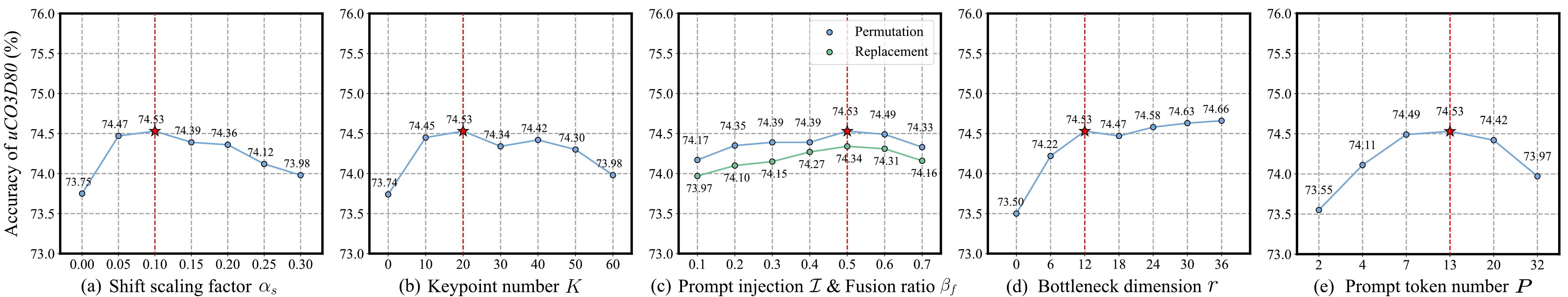}
\caption{{Ablation results on hyperparameter settings. The settings highlighted by red dashed lines denote those used in our proposed approach.}}
\label{fig-ablation-hyperparamters}
\vspace{-6pt}
\end{figure*}

\begin{table*}[h] 
\centering
\caption{{Comparison between the conference-version architecture and the proposed tpami extension. The \textcolor{blue}{blue subscripts} denote the performance improvements over the corresponding conference-version counterparts.}}
\label{tab-comparison-with-conference}
\setlength{\tabcolsep}{6pt}
\vspace{-4pt}
\begin{scriptsize}
\resizebox{1.0\linewidth}{!}{
\begin{tabular}{c|c|c|c|c|cccc}
\toprule 
\textbf{Version}&\textbf{Point Prompt}&\textbf{Keypoint Prompter}&\textbf{Point Shift Prompter}&\textbf{Prompt Propagation}&\textbf{\textit{ScanObjectNN}}&\textbf{\textit{ModelNet40}}&\textbf{\textit{GSModel60}}&\textbf{\textit{uCO3D80}} \\ 
\cmidrule(r){1-1} \cmidrule(lr){2-2} \cmidrule(lr){3-3} \cmidrule(lr){4-4} \cmidrule(l){5-9}
% \multicolumn{4}{c|}{Linear Probing of Point-PQAE\cite{pointpqae} (Baseline)}&83.19&63.78 \\
% \midrule
\textit{Conference}&\checkmark & - & - & - &84.91& 92.2 & 87.24 &65.39  \\ 
\textit{This Paper}&- & \checkmark & - & - &85.83\rlap{\textcolor{blue}{$_{0.92\uparrow}$}}& 92.7\rlap{\textcolor{blue}{$_{0.5\uparrow}$}} & 88.15\rlap{\textcolor{blue}{$_{0.91\uparrow}$}} &66.42\rlap{\textcolor{blue}{$_{1.03\uparrow}$}} \\
\midrule
\textit{Conference}&- & - & single-granular & - & 85.72 & 92.2 & 88.02 & 64.89   \\  
\textit{This Paper}&- & - & multi-granular & - & 86.55\rlap{\textcolor{blue}{$_{0.83\uparrow}$}} & 92.8\rlap{\textcolor{blue}{$_{0.6\uparrow}$}} & 89.27\rlap{\textcolor{blue}{$_{1.25\uparrow}$}} & 65.21\rlap{\textcolor{blue}{$_{0.32\uparrow}$}}    \\
\midrule   
\textit{Conference}&- & - & - & single-granular & 86.53 & 92.6 & 89.56 & 68.93   \\  
\textit{This Paper}&- & - & - & multi-granular &87.96\rlap{\textcolor{blue}{$_{1.43\uparrow}$}}& 93.2\rlap{\textcolor{blue}{$_{0.6\uparrow}$}} & 90.33\rlap{\textcolor{blue}{$_{0.77\uparrow}$}} &70.65\rlap{\textcolor{blue}{$_{1.72\uparrow}$}}    \\
\midrule   
\textit{Conference}&\checkmark & - & single-granular & single-granular &89.69&94.3&91.20&73.95\\ 
% \rowcolor{gray!15}
\textit{This Paper}&- & \checkmark & multi-granular & multi-granular & \textbf{90.63}\rlap{\textcolor{blue}{$_{0.94\uparrow}$}} & \textbf{94.6}\rlap{\textcolor{blue}{$_{0.3\uparrow}$}} & \textbf{91.61}\rlap{\textcolor{blue}{$_{0.41\uparrow}$}} & \textbf{74.53}\rlap{\textcolor{blue}{$_{0.58\uparrow}$}}    \\
\bottomrule
\end{tabular}
}
\end{scriptsize}
\vspace{-8pt}
\end{table*}

\subsection{Ablation Study}
\label{sec-ablation-study}
We conduct ablation studies on \textit{PB\_T50\_RS} of \textit{ScanObjectNN} and \textit{uCO3D80} based on Point-PQAE~\cite{pointpqae} to investigate the rationalization and effectiveness of our GAPrompt++. Overall accuracy of classification is reported.

\subsubsection{Effects of main components}
As shown in Table~\ref{tab-ablation-on-main-components}, we quantify the contribution of each main component. 
Starting from a linear probing baseline where only a lightweight task head is trained, we sequentially add each module to assess its contribution. 
As observed, the inclusion of each module leads to consistent improvements. 
Introducing the Point Shift Prompter yields the first gain by enabling instance-specific point refinement and extracting multi-granular geometric cues. Adding the Keypoint Prompter further improves performance by emphasizing salient structures and fine-grained surface variations. Their combination produces complementary benefits greater than either individually. Finally, integrating Prompt Propagation spreads geometric cues across network layers, enabling the model to exploit both local and global structure.
Remarkably, the full configuration surpasses full fine-tuning, indicating that geometry-aware prompting further enhances adaptation capacity under a parameter-efficient paradigm.

\subsubsection{Analysis on prompt attention distribution}
We visualize the attention distributions induced by our fused multi-granular geometric prompts, randomly selecting instances per granularity, as shown in Fig.~\ref{fig-multi-granular-prompts}. The results reveal that different prompt granularities attend to distinct geometric cues. The fine-grained prompts concentrate on localized structures such as edges, corners, and small components, while coarse-grained prompts emphasize overall topology and global shape contours. This complementary behavior indicates that our prompting mechanism effectively forms a hierarchical geometric understanding, enabling the model to jointly capture local detail and global semantics for more robust 3D representation learning.

\subsubsection{Analysis on the shift scaling factor $\alpha_s$}
As shown in Fig.~\ref{fig-ablation-hyperparamters}(a), we evaluate different settings of the shift scaling factor $\alpha_s$, which determines the magnitude of point displacement. Setting $\alpha_s{=}0$ removes point shift prompting entirely, leading to a \textbf{0.78\%} performance degradation, confirming its necessity. The sensitivity of this hyperparameter is relatively low, and the best performance is obtained around $\alpha_s{=}0.10$, which we adopt as the default setting.

\subsubsection{Analysis on the keypoint prompt number $K$}
Unlike abstract prompt tokens, the instance-specific keypoint prompts maintain explicit geometric correspondence with the input point cloud. They guide the model toward salient local structures and fine-grained surface variations, which is particularly beneficial for inherently irregular and unstructured point cloud data. As shown in Fig.~\ref{fig-ablation-hyperparamters}(b), increasing the number of keypoint prompts $K$ initially leads to performance gains, indicating that richer local geometric cues enhance adaptation. The gains become saturated beyond a moderate value of $K$, indicating that a compact prompt set already captures the most useful geometric cues with strong discriminability.

\begin{figure*}[!ht] 
\vspace{-6pt}
\centering
\includegraphics[width=1\textwidth]{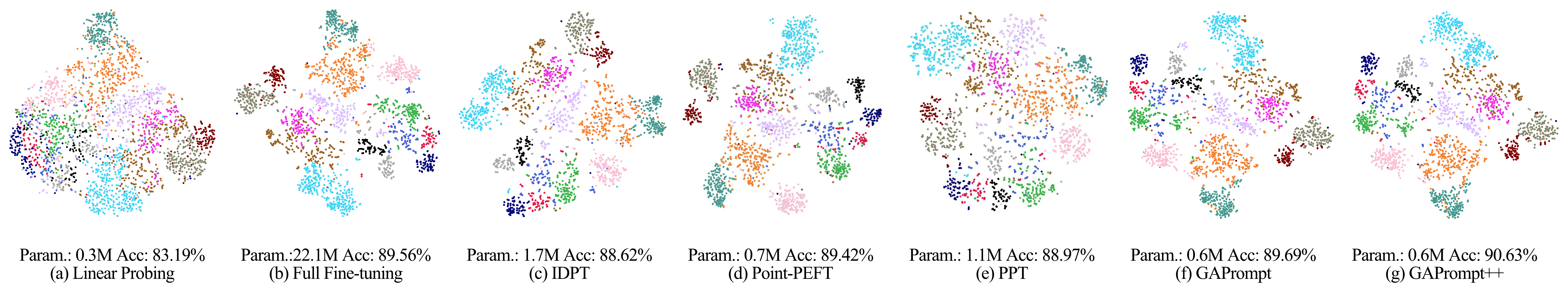}
\vspace{-20pt}
\caption{The t-SNE visualizations of test features on the \textit{PB\_T50\_RS} split of \textit{ScanObjectNN}. The embeddings are obtained from the last block of a pre-trained Point-PQAE~\cite{pointpqae} under different parameter-efficient adaptation strategies.}
\label{fig-tsne}
\end{figure*}

\begin{figure*}[!ht]
\vspace{-6pt}
\centering
\includegraphics[width=1\linewidth]{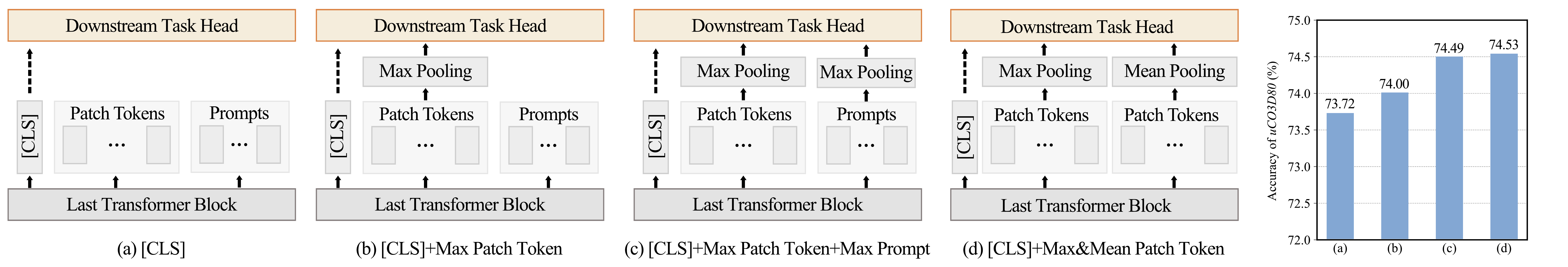}
\vspace{-20pt}
\caption{Ablation study analyzing the impact of different feature inputs fed into the downstream head and their contribution to performance.}
\label{ablation-downstream-head}
\vspace{-10pt}
\end{figure*}

\subsubsection{Analysis on prompt injection $\mathcal{I}$ and fusion ratio ${\beta}_f$}
As shown in Fig.~\ref{fig-ablation-hyperparamters}(c), we analyze two aspects of the multi-granular prompting mechanism, the manner of prompt injection $\mathcal{I}$ and the geometric prompt fusion ratio ${\beta}_f$. We evaluate two injection strategies, including direct replacement and indirect permutation of token positions. We further sweep different fusion ratios ${\beta}_f$ under both settings to examine their interaction. The results indicate that using permutation yields consistently higher performance than direct replacement, and performance peaks when ${\beta}_f$ is set to 0.5. This configuration introduces controlled stochasticity during integration, which appears to enhance robustness and stabilize optimization.

\subsubsection{Analysis on bottleneck dimension $r$}
In Prompt Propagation, a lightweight non-linear mapping is introduced to alleviate feature over-smoothing induced by spatial propagation. We therefore examine the influence of the bottleneck dimension $r$, as shown in Fig.~\ref{fig-ablation-hyperparamters}(d). The results show that setting $r=12$ offers the best balance between efficiency and accuracy, achieving a peak score of $74.53$. Although continually increasing the dimension yields marginal improvement in isolated cases, the gain is negligible relative to the additional computational cost. Hence, $r=12$ is selected as the optimal bottleneck dimension in our method.

{
{
\subsubsection{Analysis on prompt token number $P$}
As shown in Fig.~\ref{fig-ablation-hyperparamters}(e), we analyze the effect of the learnable prompt token number $P$. When $P$ is too small, the limited prompt capacity restricts task-specific adaptation, leading to suboptimal performance. Increasing $P$ brings steady gains and reaches the best performance at $P{=}13$. After that, the accuracy remains relatively stable within a moderate range and only decreases when excessive prompt tokens are introduced, likely due to redundant prompt information. These results suggest that GAPrompt++ is robust to the choice of $P$ within a reasonable range and supports strong performance across practical prompt-token settings.
}
}

\begin{table}[!t] 
\centering
\caption{Ablation on insertion layers and granularity schedule of the multi-granular prompt fusion strategy.}
\label{tab-ablation-insertion-layer}
\vspace{-4pt}
\resizebox{\linewidth}{!}{
\setlength{\tabcolsep}{1.5pt}
\begin{scriptsize}
\begin{minipage}[t]{0.54\linewidth}
\begin{tabular}{c|cc}
\toprule 
\textbf{Insertion Layers}&\textbf{\textit{ScanObjectNN}}&\textbf{\textit{uCO3D80}} \\ 
\cmidrule(r){1-1} \cmidrule(l){2-3}
1$\rightarrow$4& 90.22 & 73.98 \\ 
\textbf{1$\rightarrow$6}&\textbf{90.63}&\textbf{74.53} \\ 
1$\rightarrow$8& 90.43&74.36 \\ 
6$\rightarrow$12& 89.95&73.87 \\  
1$\rightarrow$12& 90.51&74.28 \\    
\bottomrule
\end{tabular}
\end{minipage}
\hfill

\begin{minipage}[t]{0.54\linewidth}
\begin{tabular}{c|cc}
\toprule 
\textbf{Granularities} $\mathcal{G}$&\textbf{\textit{ScanObjectNN}}&\textbf{\textit{uCO3D80}} \\ 
\cmidrule(r){1-1} \cmidrule(l){2-3}
(4, 1)& 89.67&73.59 \\ 
(4, 2, 1)& 90.45& 74.49  \\    
\textbf{(8, 4, 1)}& \textbf{90.63}&\textbf{74.53} \\ 
(16, 8, 2)& 90.24&74.08 \\ 
(16, 8, 4, 2)& 89.22&73.21 \\ 
\bottomrule
\end{tabular}
\end{minipage}
\end{scriptsize}
}
\vspace{-10pt}
\end{table}

{
{
\subsubsection{Ablation with the conference version}
To quantify the contribution of the extension, we conduct a component-wise comparison with the conference-version architecture in Table~\ref{tab-comparison-with-conference}. Specifically, we replace the original Point Prompt, single-granular Point Shift Prompter, and single-granular Prompt Propagation with the Keypoint Prompter and their multi-granular counterparts.
The results show that each extension consistently improves over its conference-version counterpart. Replacing Point Prompt with Keypoint Prompter improves performance on all datasets. The multi-granular Point Shift Prompter also brings clear gains, especially on \textit{GSModel60}, where the improvement reaches 1.25\%. The effect is more evident for Prompt Propagation, where the multi-granular version improves \textit{ScanObjectNN} and \textit{uCO3D80} by 1.43\% and 1.72\%, respectively.
These gains of GAPrompt++ are driven by the proposed multi-granular geometric prompting paradigm and its stronger geometric modeling capacity.
}
}

\subsubsection{The t-SNE visualization of instance embeddings}
Fig.~\ref{fig-tsne} presents t-SNE~\cite{tsne} embeddings for linear probing, full fine-tuning, IDPT, Point-PEFT, PPT, our GAPrompt, and GAPrompt++ on the \textit{PB\_T50\_RS} split of \textit{ScanObjectNN}. As shown in Fig.~\ref{fig-tsne}(a), features from the pre-trained model exhibit weak class separation due to the large domain gap between synthetic \textit{ShapeNet} pre-training data and real-world scans, highlighting the need for downstream adaptation. With full fine-tuning (Fig.~\ref{fig-tsne}(b)), class clusters become clearer as all parameters are optimized. Fig.~\ref{fig-tsne}(c–g) further shows that GAPrompt++ yields more compact and separable clusters than other PEFT methods, while using significantly fewer trainable parameters and achieving higher recognition accuracy.

\subsubsection{Ablation on downstream head input}
Fig.~\ref{ablation-downstream-head} analyzes the effect of different feature inputs to the downstream head, including the [CLS] token, max patch token, max prompt token, and mean patch token.
Empirically, the best performance arises from combining the [CLS] token, max patch token, and mean patch token (Fig.~\ref{ablation-downstream-head}(d)). The [CLS] and max patch tokens convey discriminative knowledge extracted by the pre-trained model, while the mean patch token supplies global geometric cues.

\subsubsection{Ablation on multi-granular prompt fusion strategy}
To evaluate the effectiveness of injecting multi-granular geometric prompts across the network hierarchy, we ablate on the prompt insertion layers and granularity schedule $\mathcal{G}$. As shown in {{Table}~\ref{tab-ablation-insertion-layer}}, inserting prompts from shallow to mid-level layers (1→6) achieves the best performance, indicating that early geometric cues can be progressively refined through subsequent layers while avoiding over-saturation in deeper blocks.
Similarly, the granularity schedule (8, 4, 1) yields the highest accuracy, suggesting that a compact set of complementary scales provides sufficient geometric diversity while keeping optimization stable. Extremely sparse or overly dense granularity configurations lead to inferior results, highlighting the importance of a balanced hierarchical prompt design. Together, these results validate the proposed fusion strategy and its role in effectively propagating multi-granular geometry through the network.

{
{
\subsubsection{Efficiency Analysis}
To better assess the practical cost of GAPrompt++, we compare different PEFT methods in terms of trainable parameters, FLOPs, inference latency, and memory footprint. For \textit{ScanObjectNN} classification with Point-PQAE~\cite{pointpqae}, all methods are evaluated with batch size 32 and 8 data-loading workers, and we report the average inference latency over the full test set together with the peak GPU memory. For \textit{S3DIS} scene segmentation with Utonia~\cite{utonia}, we use batch size 1 and load one scene with 100K points at each iteration.
Under the Point-PQAE setting, GAPrompt++ remains highly lightweight with 0.6M trainable parameters and \textbf{5.0G} FLOPs and the latency is only \textbf{8.32} ms, corresponding to a throughput of over \textbf{100} FPS. 
Under the Utonia setting, GAPrompt++ uses only 1.0M trainable parameters, with \textbf{51.3G} FLOPs, \textbf{264.97} ms latency, and \textbf{14.2} GB memory. This corresponds to nearly \textbf{4} FPS when processing large-scale scenes with 100K points. Its overhead over GAPrompt remains small, while the overall efficiency is stronger than other methods~\cite{wang2025pointlora,liang2025pointgst,gem}. These results show that the proposed multi-granular prompting design maintains favorable efficiency in both object-level classification and large-scale scene segmentation settings.
}
\vspace{-8pt}
}

\begin{table}[!t]
\centering
\caption{{Efficiency comparison across methods on trainable parameters, computational cost, latency, and memory footprints.}}
\label{tab-ablation-efficiency}
\vspace{-4pt}
\setlength{\extrarowheight}{1pt}
\begin{small}
\resizebox{\linewidth}{!}{
\begin{tabular}{l|ccccc}
\toprule
\textbf{Methods} & \textbf{Param.(M)} & \textbf{FLOPs(G)} & \textbf{Latency(ms)} & \textbf{Memory(GB)} \\
\cmidrule(r){1-1}\cmidrule(l){2-5}
Point-PQAE\cite{pointpqae} & 22.1 & 4.8 & 7.20 & 17.2 \\ 
\cmidrule(r){1-1}\cmidrule(l){2-5}
w.IDPT\cite{idpt} &1.7&7.2&7.62&19.9\\
w.Point-PEFT\cite{tang2024Point-PEFT} &0.7&7.0&10.33&19.8 \\ 
w.PPT\cite{zhang2024ppt} &1.1&10.8&17.85&21.3 \\
w.PointGST\cite{liang2025pointgst} &0.6&4.8&7.94&17.7 \\
w.PointLoRA\cite{wang2025pointlora} &0.8&8.7&15.09&19.4 \\
\rowcolor{gray!15}
w.GAPrompt\cite{ai2025gaprompt} &0.6&5.0&8.15&17.8 \\
\rowcolor{gray!15}
w.\textbf{GAPrompt++} &0.6&5.0&8.32&17.5 \\
\midrule
Utonia\cite{utonia} & 157.7 & 50.2 & 211.15 & 14.1 \\ 
\cmidrule(r){1-1}\cmidrule(l){2-5}
w.PointLoRA\cite{wang2025pointlora} &2.1&55.4&487.91&15.2 \\
w.PointGST\cite{liang2025pointgst} &1.2&52.6&1173.45&14.4 \\
w.GEM\cite{gem} &1.9&53.4&291.79&14.6 \\
\rowcolor{gray!15}
w.GAPrompt\cite{ai2025gaprompt} &1.0&51.1&257.85&14.2 \\
\rowcolor{gray!15}
w.\textbf{GAPrompt++} &1.0&51.3&264.97&14.2 \\
\bottomrule
\end{tabular}
}
\end{small}
\vspace{-8pt}
\end{table}

{
{
\vspace{-12pt}
\subsection{Experiments on Outdoor 3D Object Detection}
 To examine the generalizability of GAPrompt++ beyond classification and segmentation, we further conduct outdoor 3D object detection experiments on the large-scale \textit{NuScenes} autonomous driving benchmark~\cite{nuscenes} with Utonia~\cite{utonia} as backbone and FCAF3D~\cite{fcaf3d} as detector. Following the standard \textit{NuScenes} evaluation protocol, we report the mean Average Precision (mAP) and the nuScenes Detection Score (NDS) to measure detection performance. Detection is a fundamental perception component in embodied and autonomous driving systems, as it requires localizing object instances from large-scale outdoor point clouds before downstream planning or interaction. Therefore, this setting provides a practical and task-oriented testbed for evaluating the transferability of our geometry-aware prompting framework.
}

{
 As shown in Table~\ref{tab-3d-detection}, GAPrompt++ achieves the best performance among parameter-efficient tuning methods, reaching \textbf{66.8} mAP and \textbf{70.6} NDS with only \textbf{15.5M} trainable parameters. Compared with GAPrompt~\cite{ai2025gaprompt}, the proposed extension improves detection accuracy by \textbf{+1.9} mAP and \textbf{+2.4} NDS, and it also surpasses strong PEFT baselines such as PointGST~\cite{liang2025pointgst} by \textbf{+1.7} mAP and \textbf{+2.1} NDS. GAPrompt++ also remains close to the full fine-tuning reference while using a much smaller adaptation budget. These results suggest that our method can scale to broader large-scene 3D perception tasks in real-world autonomous driving scenarios.
}
}

\begin{table}[!t]
\centering
\caption{{Experiments on 3D object detection on the \textit{NuScenes} benchmark for autonomous driving scenarios.}}
\label{tab-3d-detection}
\setlength{\extrarowheight}{-1pt}
\setlength{\tabcolsep}{8pt}
\vspace{-4pt}
\begin{small}
\resizebox{\linewidth}{!}{
\begin{tabular}{llcccc}
\toprule
&\textbf{Methods} & \textbf{Publication} & \textbf{Param.(M)} & \textbf{mAP} & \textbf{NDS} \\
\midrule
\midrule
&\multicolumn{1}{|l}{Full Fine-Tuning} & - & 151.7 & 67.0 & 71.5 \\
\cmidrule{2-6}
\multirow{10}{*}{\rotatebox{90}{Utonia\cite{utonia}+FCAF3D\cite{fcaf3d}}}&\multicolumn{5}{|c}{\textit{Reparameterization-based Methods}}  \\ 
\cmidrule{2-6}
&\multicolumn{1}{|l}{PointLoRA~\cite{wang2025pointlora}} & \textit{CVPR 25} & 17.6 & 64.4 & 66.0 \\
&\multicolumn{1}{|l}{MoST~\cite{han2025most}} & \textit{CVPR 25} &17.2&64.8&66.5\\
\cmidrule{2-6}
&\multicolumn{5}{|c}{\textit{Adapter-based Methods}}  \\ 
\cmidrule{2-6}
&\multicolumn{1}{|l}{DAPT~\cite{zhou2024DAPT}} & CVPR 24 & 16.8 & 64.5 & 67.8 \\
&\multicolumn{1}{|l}{PointGST~\cite{liang2025pointgst}} & \textit{TPAMI 25} & 16.1 & 65.1 & 68.5 \\
% &\multicolumn{1}{|l}{PMA} & \textit{CVPR 25} & 5.64 & 84.00 & 86.1 \\
\cmidrule{2-6}
&\multicolumn{5}{|c}{\textit{Prompt-based Methods}}  \\ 
\cmidrule{2-6}
&\multicolumn{1}{|l}{Point-PEFT~\cite{tang2024Point-PEFT}} & \textit{AAAI 24} & 17.7 & 64.3 & 67.4 \\
&\multicolumn{1}{|l}{PPT~\cite{zhang2024ppt}} & \textit{MM 25} & 19.0 & 63.1 & 66.5 \\
\rowcolor{gray!15}
\cellcolor{white}
&\multicolumn{1}{|l}{GAPrompt~\cite{ai2025gaprompt}} & \textit{ICML 25} & 15.5 & 64.9 & 68.2 \\
\rowcolor{gray!15}
\cellcolor{white}
&\multicolumn{1}{|l}{\textbf{GAPrompt++}}&\textit{This Paper}&\textbf{15.5}&\textbf{{66.8}}& \textbf{70.6} \\
 \bottomrule
\end{tabular}
}
\end{small}
\vspace{-8pt}
\end{table}

\begin{table}[!t]
\centering
\caption{Training details of downstream adaptation datasets.}
\vspace{-8pt}
\begin{small}
{\scriptsize (a) Object-level datasets.\par}
\vspace{3pt}
\resizebox{\linewidth}{!}{
\begin{tabular}{l|ccccc}
\toprule
\textbf{{Dataset}} &\textit{\textbf{ScanObjectNN}}&{\textit{\textbf{ModelNet40}}}&{\textit{\textbf{ShapeNetPart}}}&{\textit{\textbf{GSModel60}}}&{\textit{\textbf{uCO3D80}}} \\ 
\cmidrule(r){1-1}\cmidrule(l){2-6}
Optimizer & AdamW & AdamW & AdamW & AdamW & AdamW\\ 
Learning rate & 0.0005 & 0.0005 & 0.0002 & 0.0005 & 0.0005\\ 
Weight decay & 0.05 & 0.05 & 0.05 & 0.05 & 0.05 \\ 
Scheduler & cosine & cosine & cosine & cosine & cosine \\ 
Training epoch & 400 & 400 & 400 & 400 & 400 \\ 
Warmup epoch & 10 & 10 & 10 & 10 & 10 \\ 
Batch size & 32 & 32 & 32 & 32 & 32 \\ 
Point number & 2048 & 1024 & 2048 & 2048 & 2048 \\  
Patch number & 128 & 64 & 128 & 128 & 128 \\ 
Patch size & 32 & 32 & 32 & 32 & 32 \\  
\bottomrule
\end{tabular}
}

\vspace{4pt}
{\scriptsize (b) Scene-level datasets.\par}
\vspace{2pt}
\setlength{\tabcolsep}{6pt}
\resizebox{0.9\linewidth}{!}{
\begin{tabular}{l|cccc}
\toprule
\textbf{Dataset} & \textit{\textbf{~~~S3DIS~~~}} & \textit{\textbf{~~~ScanNet~~~}} & \textit{\textbf{NuScenes Seg.}} & \textit{\textbf{NuScenes Det.}} \\ 
\cmidrule(r){1-1}\cmidrule(l){2-5}
Optimizer & AdamW & AdamW & AdamW & AdamW \\ 
Learning rate & 0.002 & 0.002 & 0.002 & 0.0005 \\ 
Weight decay & 0.02 & 0.02 & 0.02 & 0.0001  \\ 
Scheduler & cosine & cosine & cosine & cosine  \\ 
Training epoch & 100 & 100 & 100 & 24  \\ 
Warmup epoch & 2 & 2 & 2 & 1  \\ 
Batch size & 4 & 4 & 4 & 4  \\ 
Voxel size & 0.01m & 0.01m & 0.01m & 0.075m  \\  
Patch size & 1024 & 1024 & 1024 & 1024  \\  
\bottomrule
\end{tabular}
}
\end{small}
\label{tab-train-details}
\vspace{-10pt}
\end{table}

\subsection{Training Detail}
 We adopt downstream adaptation configurations in alignment with the pioneering work Point-MAE~\cite{pang2022PointMAE} for object-level experiments and Utonia~\cite{utonia} for the outdoor detection setting. The detailed configurations are provided in Table \ref{tab-train-details}. Given that Point-FEMAE~\cite{zha2023PointFEMAE} and PointPQAE~\cite{pointpqae} extend Point-MAE with several additional modules, we follow the approach of DAPT~\cite{zhou2024DAPT} by only loading pre-trained weights into a Point-MAE model for efficient fine-tuning, while excluding the residual components of ReCon and Point-FEMAE, which also reduces the computation overhead in Table~\ref{tab-finetuning}. For scene-level segmentation, both Concerto and Utonia are adapted using Point Transformer v3 for adaptation, while the outdoor detection setting follows the Utonia+FCAF3D configuration. All experiments are conducted on GeForce RTX 5090 cards using PyTorch 2.8.0.

\section{Conclusion}
In this paper, we present GAPrompt++, a multi-granular geometry-aware prompting framework for parameter-efficient adaptation of pre-trained 3D vision models. We demonstrate that explicitly modeling instance-specific geometric cues is key to strengthening geometric awareness within prompting. To this end, we further advance point-level prompting with hierarchical geometric priors and an improved propagation mechanism that jointly enrich prompt expressiveness and structural alignment. 
Extensive experiments across synthetic and real-world benchmarks, including two newly curated datasets, validate the effectiveness and generality of our framework. Moreover, GAPrompt++ demonstrates cross-modal adaptability, providing a practical pathway to leverage pre-trained models from text and 2D vision domains for 3D understanding while reducing reliance on large-scale point cloud supervision.

\section{Acknowledgment}
This work was supported in part by the National Key Research and Development Program of China under Grant 2022ZD0114903, and in part by the National Natural Science Foundation of China (62376011), and the National Key R\&D Program of China (2024YFA1410000).


% \section{References Section}
% You can use a bibliography generated by BibTeX as a .bbl file.
% The IEEEtran BibTeX style support page is:
 

% \newpage
\vspace{-40pt}


\begin{thebibliography}{10}
\providecommand{\url}[1]{#1}
\csname url@samestyle\endcsname
\providecommand{\newblock}{\relax}
\providecommand{\bibinfo}[2]{#2}
\providecommand{\BIBentrySTDinterwordspacing}{\spaceskip=0pt\relax}
\providecommand{\BIBentryALTinterwordstretchfactor}{4}
\providecommand{\BIBentryALTinterwordspacing}{\spaceskip=\fontdimen2\font plus
\BIBentryALTinterwordstretchfactor\fontdimen3\font minus \fontdimen4\font\relax}
\providecommand{\BIBforeignlanguage}[2]{{%
\expandafter\ifx\csname l@#1\endcsname\relax
\typeout{** WARNING: IEEEtran.bst: No hyphenation pattern has been}%
\typeout{** loaded for the language `#1'. Using the pattern for}%
\typeout{** the default language instead.}%
\else
\language=\csname l@#1\endcsname
\fi
#2}}
\providecommand{\BIBdecl}{\relax}
\BIBdecl

\bibitem{mvsnet}
Y.~Yao, Z.~Luo, S.~Li, T.~Fang, and L.~Quan, ``Mvsnet: Depth inference for unstructured multi-view stereo,'' in \emph{Proceedings of the European conference on computer vision (ECCV)}, 2018, pp. 767--783.

\bibitem{nerf}
B.~Mildenhall, P.~P. Srinivasan, M.~Tancik, J.~T. Barron, R.~Ramamoorthi, and R.~Ng, ``Nerf: Representing scenes as neural radiance fields for view synthesis,'' in \emph{European Conference on Computer Vision}.\hskip 1em plus 0.5em minus 0.4em\relax Springer, 2020, pp. 405--421.

\bibitem{gaussiansplatting}
B.~Kerbl, G.~Kopanas, T.~Leimk{\"u}hler, and G.~Drettakis, ``3d gaussian splatting for real-time radiance field rendering,'' \emph{ACM Transactions on Graphics}, vol.~42, no.~4, July 2023.

\bibitem{yu2022PointBERT}
X.~Yu, L.~Tang, Y.~Rao, T.~Huang, J.~Zhou, and J.~Lu, ``Point-bert: Pre-training 3d point cloud transformers with masked point modeling,'' in \emph{Proceedings of the IEEE/CVF conference on computer vision and pattern recognition}, 2022, pp. 19\,313--19\,322.

\bibitem{zhang2022PointM2AE}
R.~Zhang, Z.~Guo, P.~Gao, R.~Fang, B.~Zhao, D.~Wang, Y.~Qiao, and H.~Li, ``Point-m2ae: multi-scale masked autoencoders for hierarchical point cloud pre-training,'' \emph{Advances in neural information processing systems}, vol.~35, pp. 27\,061--27\,074, 2022.

\bibitem{zha2023PointFEMAE}
Y.~Zha, H.~Ji, J.~Li, R.~Li, T.~Dai, B.~Chen, Z.~Wang, and S.-T. Xia, ``Towards compact 3d representations via point feature enhancement masked autoencoders,'' in \emph{Proceedings of the AAAI Conference on Artificial Intelligence}, vol.~38, 2024, pp. 6962--6970.

\bibitem{lester2021power}
B.~Lester, R.~Al-Rfou, and N.~Constant, ``The power of scale for parameter-efficient prompt tuning,'' \emph{arXiv preprint arXiv:2104.08691}, 2021.

\bibitem{li2021prefix}
X.~L. Li and P.~Liang, ``Prefix-tuning: Optimizing continuous prompts for generation,'' \emph{arXiv preprint arXiv:2101.00190}, 2021.

\bibitem{liu2024insvp}
Z.~Liu, Y.~Peng, and J.~Zhou, ``Insvp: Efficient instance visual prompting from image itself,'' in \emph{Proceedings of the 32nd ACM International Conference on Multimedia}, 2024, pp. 6443--6452.

\bibitem{adapter}
N.~Houlsby, A.~Giurgiu, S.~Jastrzebski, B.~Morrone, Q.~De~Laroussilhe, A.~Gesmundo, M.~Attariyan, and S.~Gelly, ``Parameter-efficient transfer learning for nlp,'' in \emph{International conference on machine learning}.\hskip 1em plus 0.5em minus 0.4em\relax PMLR, 2019, pp. 2790--2799.

\bibitem{prompt}
J.~He, C.~Zhou, X.~Ma, T.~Berg-Kirkpatrick, and G.~Neubig, ``Towards a unified view of parameter-efficient transfer learning,'' \emph{arXiv preprint arXiv:2110.04366}, 2021.

\bibitem{vpt}
M.~Jia, L.~Tang, B.-C. Chen, C.~Cardie, S.~Belongie, B.~Hariharan, and S.-N. Lim, ``Visual prompt tuning,'' in \emph{European Conference on Computer Vision}.\hskip 1em plus 0.5em minus 0.4em\relax Springer, 2022, pp. 709--727.

\bibitem{idpt}
Y.~Zha, J.~Wang, T.~Dai, B.~Chen, Z.~Wang, and S.-T. Xia, ``Instance-aware dynamic prompt tuning for pre-trained point cloud models,'' in \emph{Proceedings of the IEEE/CVF International Conference on Computer Vision}, 2023, pp. 14\,161--14\,170.

\bibitem{zhou2024DAPT}
X.~Zhou, D.~Liang, W.~Xu, X.~Zhu, Y.~Xu, Z.~Zou, and X.~Bai, ``Dynamic adapter meets prompt tuning: Parameter-efficient transfer learning for point cloud analysis,'' in \emph{Proceedings of the IEEE/CVF Conference on Computer Vision and Pattern Recognition}, 2024, pp. 14\,707--14\,717.

\bibitem{ai2025gaprompt}
Z.~Ai, Z.~Liu, Y.~Lei, Z.~Cui, X.~Zou, and J.~Zhou, ``Gaprompt: Geometry-aware point cloud prompt for 3d vision model,'' in \emph{International Conference on Machine Learning}, 2025.

\bibitem{rt2}
B.~Zitkovich, T.~Yu, S.~Xu, P.~Xu, T.~Xiao, F.~Xia, J.~Wu, P.~Wohlhart, S.~Welker, A.~Wahid \emph{et~al.}, ``Rt-2: Vision-language-action models transfer web knowledge to robotic control,'' in \emph{Conference on Robot Learning}.\hskip 1em plus 0.5em minus 0.4em\relax PMLR, 2023, pp. 2165--2183.

\bibitem{openvla}
M.~J. Kim, K.~Pertsch, S.~Karamcheti, T.~Xiao, A.~Balakrishna, S.~Nair, R.~Rafailov, E.~P. Foster, P.~R. Sanketi, Q.~Vuong \emph{et~al.}, ``Openvla: An open-source vision-language-action model,'' in \emph{Conference on Robot Learning}.\hskip 1em plus 0.5em minus 0.4em\relax PMLR, 2025, pp. 2679--2713.

\bibitem{asap}
T.~He, J.~Gao, W.~Xiao, Y.~Zhang, Z.~Wang, J.~Wang, Z.~Luo, G.~He, N.~Sobanbab, C.~Pan \emph{et~al.}, ``Asap: Aligning simulation and real-world physics for learning agile humanoid whole-body skills,'' \emph{arXiv preprint arXiv:2502.01143}, 2025.

\bibitem{navgpt}
G.~Zhou, Y.~Hong, and Q.~Wu, ``Navgpt: Explicit reasoning in vision-and-language navigation with large language models,'' in \emph{Proceedings of the AAAI Conference on Artificial Intelligence}, vol.~38, no.~7, 2024, pp. 7641--7649.

\bibitem{objectnav}
D.~S. Chaplot, D.~P. Gandhi, A.~Gupta, and R.~R. Salakhutdinov, ``Object goal navigation using goal-oriented semantic exploration,'' \emph{Advances in Neural Information Processing Systems}, vol.~33, pp. 4247--4258, 2020.

\bibitem{pointnet}
C.~R. Qi, H.~Su, K.~Mo, and L.~J. Guibas, ``Pointnet: Deep learning on point sets for 3d classification and segmentation,'' in \emph{Proceedings of the IEEE conference on computer vision and pattern recognition}, 2017, pp. 652--660.

\bibitem{pointnet++}
C.~R. Qi, L.~Yi, H.~Su, and L.~J. Guibas, ``Pointnet++: Deep hierarchical feature learning on point sets in a metric space,'' \emph{Advances in neural information processing systems}, vol.~30, 2017.

\bibitem{dgcnn}
A.~V. Phan, M.~Le~Nguyen, Y.~L.~H. Nguyen, and L.~T. Bui, ``Dgcnn: A convolutional neural network over large-scale labeled graphs,'' \emph{Neural Networks}, vol. 108, pp. 533--543, 2018.

\bibitem{pointnext}
G.~Qian, Y.~Li, H.~Peng, J.~Mai, H.~Hammoud, M.~Elhoseiny, and B.~Ghanem, ``Pointnext: Revisiting pointnet++ with improved training and scaling strategies,'' \emph{Advances in neural information processing systems}, vol.~35, pp. 23\,192--23\,204, 2022.

\bibitem{dosovitskiy2020vit}
A.~Dosovitskiy, L.~Beyer, A.~Kolesnikov, D.~Weissenborn, X.~Zhai, T.~Unterthiner, M.~Dehghani, M.~Minderer, G.~Heigold, S.~Gelly, J.~Uszkoreit, and N.~Houlsby, ``An image is worth 16x16 words: Transformers for image recognition at scale,'' \emph{ICLR}, 2021.

\bibitem{visionmamba}
L.~Zhu, B.~Liao, Q.~Zhang, X.~Wang, W.~Liu, and X.~Wang, ``Vision mamba: efficient visual representation learning with bidirectional state space model,'' in \emph{Proceedings of the 41st International Conference on Machine Learning}, 2024, pp. 62\,429--62\,442.

\bibitem{pointcontrast}
S.~Xie, J.~Gu, D.~Guo, C.~R. Qi, L.~Guibas, and O.~Litany, ``Pointcontrast: Unsupervised pre-training for 3d point cloud understanding,'' in \emph{European conference on computer vision}.\hskip 1em plus 0.5em minus 0.4em\relax Springer, 2020, pp. 574--591.

\bibitem{clip2point}
T.~Huang, B.~Dong, Y.~Yang, X.~Huang, R.~W. Lau, W.~Ouyang, and W.~Zuo, ``Clip2point: Transfer clip to point cloud classification with image-depth pre-training,'' in \emph{Proceedings of the IEEE/CVF International Conference on Computer Vision (ICCV)}, October 2023, pp. 22\,157--22\,167.

\bibitem{pointclipv2}
X.~Zhu, R.~Zhang, B.~He, Z.~Guo, Z.~Zeng, Z.~Qin, S.~Zhang, and P.~Gao, ``Pointclip v2: Prompting clip and gpt for powerful 3d open-world learning,'' in \emph{Proceedings of the IEEE/CVF International Conference on Computer Vision}, 2023, pp. 2639--2650.

\bibitem{ACT_rotate}
R.~Dong, Z.~Qi, L.~Zhang, J.~Zhang, J.~Sun, Z.~Ge, L.~Yi, and K.~Ma, ``Autoencoders as cross-modal teachers: Can pretrained 2d image transformers help 3d representation learning?'' \emph{arXiv preprint arXiv:2212.08320}, 2022.

\bibitem{qi2023ReCon}
Z.~Qi, R.~Dong, G.~Fan, Z.~Ge, X.~Zhang, K.~Ma, and L.~Yi, ``Contrast with reconstruct: Contrastive 3d representation learning guided by generative pretraining,'' in \emph{International Conference on Machine Learning}.\hskip 1em plus 0.5em minus 0.4em\relax PMLR, 2023, pp. 28\,223--28\,243.

\bibitem{recon++}
Z.~Qi, R.~Dong, S.~Zhang, H.~Geng, C.~Han, Z.~Ge, L.~Yi, and K.~Ma, ``Shapellm: Universal 3d object understanding for embodied interaction,'' \emph{arXiv preprint arXiv:2402.17766}, 2024.

\bibitem{clip}
A.~Radford, J.~W. Kim, C.~Hallacy, A.~Ramesh, G.~Goh, S.~Agarwal, G.~Sastry, A.~Askell, P.~Mishkin, J.~Clark \emph{et~al.}, ``Learning transferable visual models from natural language supervision,'' in \emph{International conference on machine learning}.\hskip 1em plus 0.5em minus 0.4em\relax PmLR, 2021, pp. 8748--8763.

\bibitem{dinov3}
O.~Sim{\'e}oni, H.~V. Vo, M.~Seitzer, F.~Baldassarre, M.~Oquab, C.~Jose, V.~Khalidov, M.~Szafraniec, S.~Yi, M.~Ramamonjisoa \emph{et~al.}, ``Dinov3,'' \emph{arXiv preprint arXiv:2508.10104}, 2025.

\bibitem{bert}
J.~Devlin, M.-W. Chang, K.~Lee, and K.~Toutanova, ``Bert: Pre-training of deep bidirectional transformers for language understanding,'' in \emph{Proceedings of the 2019 conference of the North American chapter of the association for computational linguistics: human language technologies, volume 1 (long and short papers)}, 2019, pp. 4171--4186.

\bibitem{mae}
K.~He, X.~Chen, S.~Xie, Y.~Li, P.~Doll{\'a}r, and R.~Girshick, ``Masked autoencoders are scalable vision learners,'' in \emph{Proceedings of the IEEE/CVF conference on computer vision and pattern recognition}, 2022, pp. 16\,000--16\,009.

\bibitem{maskpoint}
H.~Liu, M.~Cai, and Y.~J. Lee, ``Masked discrimination for self-supervised learning on point clouds,'' in \emph{European Conference on Computer Vision}.\hskip 1em plus 0.5em minus 0.4em\relax Springer, 2022, pp. 657--675.

\bibitem{pcpmae}
X.~Zhang, S.~Zhang, and J.~Yan, ``Pcp-mae: Learning to predict centers for point masked autoencoders,'' \emph{Advances in Neural Information Processing Systems}, vol.~37, pp. 80\,303--80\,327, 2024.

\bibitem{pointpqae}
X.~Zhang and J.~Yan, ``Towards more diverse and challenging pre-training for point cloud learning: Self-supervised cross reconstruction with decoupled views,'' in \emph{Proceedings of the IEEE/CVF International Conference on Computer Vision}, 2025, pp. 28\,696--28\,706.

\bibitem{pointgpt}
G.~Chen, M.~Wang, Y.~Yang, K.~Yu, L.~Yuan, and Y.~Yue, ``Pointgpt: Auto-regressively generative pre-training from point clouds,'' \emph{Advances in Neural Information Processing Systems}, vol.~36, 2024.

\bibitem{tap}
Z.~Wang, X.~Yu, Y.~Rao, J.~Zhou, and J.~Lu, ``Take-a-photo: 3d-to-2d generative pre-training of point cloud models,'' in \emph{Proceedings of the IEEE/CVF International Conference on Computer Vision}, 2023, pp. 5640--5650.

\bibitem{i2pmae}
R.~Zhang, L.~Wang, Y.~Qiao, P.~Gao, and H.~Li, ``Learning 3d representations from 2d pre-trained models via image-to-point masked autoencoders,'' in \emph{Proceedings of the IEEE/CVF conference on computer vision and pattern recognition}, 2023, pp. 21\,769--21\,780.

\bibitem{pointsd}
Y.~Chen, S.~Zhao, L.~Duan, C.~Ding, and D.~Tao, ``Harnessing text-to-image diffusion models for point cloud self-supervised learning,'' \emph{arXiv preprint arXiv:2507.09102}, 2025.

\bibitem{dino}
M.~Caron, H.~Touvron, I.~Misra, H.~J{\'e}gou, J.~Mairal, P.~Bojanowski, and A.~Joulin, ``Emerging properties in self-supervised vision transformers,'' in \emph{Proceedings of the IEEE/CVF international conference on computer vision}, 2021, pp. 9650--9660.

\bibitem{diffusion}
R.~Rombach, A.~Blattmann, D.~Lorenz, P.~Esser, and B.~Ommer, ``High-resolution image synthesis with latent diffusion models,'' in \emph{Proceedings of the IEEE/CVF conference on computer vision and pattern recognition}, 2022, pp. 10\,684--10\,695.

\bibitem{vp}
H.~Bahng, A.~Jahanian, S.~Sankaranarayanan, and P.~Isola, ``Exploring visual prompts for adapting large-scale models,'' \emph{arXiv preprint arXiv:2203.17274}, 2022.

\bibitem{dept}
Z.~Shi and A.~Lipani, ``Dept: Decomposed prompt tuning for parameter-efficient fine-tuning,'' \emph{arXiv preprint arXiv:2309.05173}, 2023.

\bibitem{adaptformer}
S.~Chen, C.~Ge, Z.~Tong, J.~Wang, Y.~Song, J.~Wang, and P.~Luo, ``Adaptformer: Adapting vision transformers for scalable visual recognition,'' \emph{Advances in Neural Information Processing Systems}, vol.~35, pp. 16\,664--16\,678, 2022.

\bibitem{biadaptformer}
S.~Jie, H.~Wang, and Z.-H. Deng, ``Revisiting the parameter efficiency of adapters from the perspective of precision redundancy,'' in \emph{Proceedings of the IEEE/CVf international conference on computer vision}, 2023, pp. 17\,217--17\,226.

\bibitem{wang2025pointlora}
S.~Wang, X.~Liu, L.~Kong, J.~Xu, C.~Hu, G.~Fang, W.~Li, J.~Zhu, and X.~Wang, ``Pointlora: Low-rank adaptation with token selection for point cloud learning,'' in \emph{Proceedings of the Computer Vision and Pattern Recognition Conference}, 2025, pp. 6605--6615.

\bibitem{han2025most}
X.~Han, Y.~Tang, J.~Xu, and X.~Li, ``Most: Efficient monarch sparse tuning for 3d representation learning,'' in \emph{Proceedings of the Computer Vision and Pattern Recognition Conference}, 2025, pp. 6584--6594.

\bibitem{liang2025pointgst}
D.~Liang, T.~Feng, X.~Zhou, Y.~Zhang, Z.~Zou, and X.~Bai, ``Parameter-efficient fine-tuning in spectral domain for point cloud learning,'' \emph{IEEE Transactions on Pattern Analysis and Machine Intelligence}, 2025.

\bibitem{zha2025pma}
Y.~Zha, Y.~Wang, H.~Guo, J.~Wang, T.~Dai, B.~Chen, Z.~Ouyang, X.~Yuerong, K.~Chen, and S.-T. Xia, ``Pma: Towards parameter-efficient point cloud understanding via point mamba adapter,'' in \emph{Proceedings of the Computer Vision and Pattern Recognition Conference}, 2025, pp. 16\,976--16\,986.

\bibitem{gem}
L.~Tang, Z.~Chen, and D.~Tao, ``On geometry-enhanced parameter-efficient fine-tuning for 3d scene segmentation,'' \emph{Advances in Neural Information Processing Systems}, vol.~38, pp. 168\,533--168\,556, 2025.

\bibitem{tang2024Point-PEFT}
Y.~Tang, R.~Zhang, Z.~Guo, X.~Ma, B.~Zhao, Z.~Wang, D.~Wang, and X.~Li, ``Point-peft: Parameter-efficient fine-tuning for 3d pre-trained models,'' in \emph{Proceedings of the AAAI Conference on Artificial Intelligence}, vol.~38, 2024, pp. 5171--5179.

\bibitem{zhang2024ppt}
S.~Zhang, Z.~Qi, R.~Dong, X.~Bai, and X.~Wei, ``Positional prompt tuning for efficient 3d representation learning,'' in \emph{Proceedings of the ACM International Conference on Multimedia}, 2025.

\bibitem{scannet}
A.~Dai, A.~X. Chang, M.~Savva, M.~Halber, T.~Funkhouser, and M.~Nie{\ss}ner, ``Scannet: Richly-annotated 3d reconstructions of indoor scenes,'' in \emph{Proceedings of the IEEE conference on computer vision and pattern recognition}, 2017, pp. 5828--5839.

\bibitem{shapenet55}
A.~X. Chang, T.~Funkhouser, L.~Guibas, P.~Hanrahan, Q.~Huang, Z.~Li, S.~Savarese, M.~Savva, S.~Song, H.~Su \emph{et~al.}, ``Shapenet: An information-rich 3d model repository,'' \emph{arXiv preprint arXiv:1512.03012}, 2015.

\bibitem{modelnet40}
Z.~Wu, S.~Song, A.~Khosla, F.~Yu, L.~Zhang, X.~Tang, and J.~Xiao, ``3d shapenets: A deep representation for volumetric shapes,'' in \emph{Proceedings of the IEEE conference on computer vision and pattern recognition}, 2015, pp. 1912--1920.

\bibitem{omniobject3d}
T.~Wu, J.~Zhang, X.~Fu, Y.~Wang, J.~Ren, L.~Pan, W.~Wu, L.~Yang, J.~Wang, C.~Qian \emph{et~al.}, ``Omniobject3d: Large-vocabulary 3d object dataset for realistic perception, reconstruction and generation,'' in \emph{Proceedings of the IEEE/CVF Conference on Computer Vision and Pattern Recognition}, 2023, pp. 803--814.

\bibitem{objaverse}
M.~Deitke, D.~Schwenk, J.~Salvador, L.~Weihs, O.~Michel, E.~VanderBilt, L.~Schmidt, K.~Ehsani, A.~Kembhavi, and A.~Farhadi, ``Objaverse: A universe of annotated 3d objects,'' in \emph{Proceedings of the IEEE/CVF conference on computer vision and pattern recognition}, 2023, pp. 13\,142--13\,153.

\bibitem{objaversexl}
M.~Deitke, R.~Liu, M.~Wallingford, H.~Ngo, O.~Michel, A.~Kusupati, A.~Fan, C.~Laforte, V.~Voleti, S.~Y. Gadre \emph{et~al.}, ``Objaverse-xl: A universe of 10m+ 3d objects,'' \emph{Advances in Neural Information Processing Systems}, vol.~36, pp. 35\,799--35\,813, 2023.

\bibitem{scanobjectnn}
M.~A. Uy, Q.-H. Pham, B.-S. Hua, T.~Nguyen, and S.-K. Yeung, ``Revisiting point cloud classification: A new benchmark dataset and classification model on real-world data,'' in \emph{Proceedings of the IEEE/CVF international conference on computer vision}, 2019, pp. 1588--1597.

\bibitem{dropout}
N.~Srivastava, G.~Hinton, A.~Krizhevsky, I.~Sutskever, and R.~Salakhutdinov, ``Dropout: a simple way to prevent neural networks from overfitting,'' \emph{The journal of machine learning research}, vol.~15, no.~1, pp. 1929--1958, 2014.

\bibitem{s3dis}
I.~Armeni, O.~Sener, A.~R. Zamir, H.~Jiang, I.~Brilakis, M.~Fischer, and S.~Savarese, ``3d semantic parsing of large-scale indoor spaces,'' in \emph{Proceedings of the IEEE conference on computer vision and pattern recognition}, 2016, pp. 1534--1543.

\bibitem{nuscenes}
H.~Caesar, V.~Bankiti, A.~H. Lang, S.~Vora, V.~E. Liong, Q.~Xu, A.~Krishnan, Y.~Pan, G.~Baldan, and O.~Beijbom, ``nuscenes: A multimodal dataset for autonomous driving,'' in \emph{Proceedings of the IEEE/CVF conference on computer vision and pattern recognition}, 2020, pp. 11\,621--11\,631.

\bibitem{voting}
Y.~Liu, B.~Fan, S.~Xiang, and C.~Pan, ``Relation-shape convolutional neural network for point cloud analysis,'' in \emph{Proceedings of the IEEE/CVF conference on computer vision and pattern recognition}, 2019, pp. 8895--8904.

\bibitem{pointmlp}
X.~Ma, C.~Qin, H.~You, H.~Ran, and Y.~Fu, ``Rethinking network design and local geometry in point cloud: A simple residual mlp framework,'' \emph{arXiv preprint arXiv:2202.07123}, 2022.

\bibitem{repsurf}
H.~Ran, J.~Liu, and C.~Wang, ``Surface representation for point clouds,'' in \emph{Proceedings of the IEEE/CVF conference on computer vision and pattern recognition}, 2022, pp. 18\,942--18\,952.

\bibitem{pointvector}
X.~Deng, W.~Zhang, Q.~Ding, and X.~Zhang, ``Pointvector: A vector representation in point cloud analysis,'' in \emph{Proceedings of the IEEE/CVF conference on computer vision and pattern recognition}, 2023, pp. 9455--9465.

\bibitem{x3d}
S.~Sun, Y.~Rao, J.~Lu, and H.~Yan, ``X-3d: Explicit 3d structure modeling for point cloud recognition,'' in \emph{Proceedings of the IEEE/CVF Conference on Computer Vision and Pattern Recognition}, 2024, pp. 5074--5083.

\bibitem{map}
Y.~Liu and L.~Yi, ``Map: Unleashing hybrid mamba-transformer vision backbone's potential with masked autoregressive pretraining,'' in \emph{Proceedings of the Computer Vision and Pattern Recognition Conference}, 2025, pp. 9676--9685.

\bibitem{pang2022PointMAE}
Y.~Pang, W.~Wang, F.~E. Tay, W.~Liu, Y.~Tian, and L.~Yuan, ``Masked autoencoders for point cloud self-supervised learning,'' in \emph{European conference on computer vision}.\hskip 1em plus 0.5em minus 0.4em\relax Springer, 2022, pp. 604--621.

\bibitem{shapesplat}
Q.~Ma, Y.~Li, B.~Ren, N.~Sebe, E.~Konukoglu, T.~Gevers, L.~Van~Gool, and D.~P. Paudel, ``A large-scale dataset of gaussian splats and their self-supervised pretraining,'' in \emph{2025 International Conference on 3D Vision (3DV)}.\hskip 1em plus 0.5em minus 0.4em\relax IEEE, 2025, pp. 145--155.

\bibitem{macgs}
R.~Zhang, H.~Zhu, J.~Zhao, Q.~Zhang, X.~Cao, and Z.~Ma, ``Mitigating ambiguities in 3d classification with gaussian splatting,'' in \emph{Proceedings of the Computer Vision and Pattern Recognition Conference}, 2025, pp. 27\,275--27\,284.

\bibitem{uco3d}
X.~Liu, P.~Tayal, J.~Wang, J.~Zarzar, T.~Monnier, K.~Tertikas, J.~Duan, A.~Toisoul, J.~Y. Zhang, N.~Neverova \emph{et~al.}, ``Uncommon objects in 3d,'' in \emph{Proceedings of the Computer Vision and Pattern Recognition Conference}, 2025, pp. 14\,102--14\,113.

\bibitem{pointclip}
R.~Zhang, Z.~Guo, W.~Zhang, K.~Li, X.~Miao, B.~Cui, Y.~Qiao, P.~Gao, and H.~Li, ``Pointclip: Point cloud understanding by clip,'' in \emph{Proceedings of the IEEE/CVF conference on computer vision and pattern recognition}, 2022, pp. 8552--8562.

\bibitem{cg3d}
D.~Hegde, J.~M.~J. Valanarasu, and V.~Patel, ``Clip goes 3d: Leveraging prompt tuning for language grounded 3d recognition,'' in \emph{Proceedings of the IEEE/CVF International Conference on Computer Vision}, 2023, pp. 2028--2038.

\bibitem{pix4point}
G.~Qian, A.~Hamdi, X.~Zhang, and B.~Ghanem, ``Pix4point: Image pretrained standard transformers for 3d point cloud understanding,'' in \emph{2024 International Conference on 3D Vision (3DV)}.\hskip 1em plus 0.5em minus 0.4em\relax IEEE, 2024, pp. 1280--1290.

\bibitem{any2point}
Y.~Tang, R.~Zhang, J.~Liu, Z.~Guo, B.~Zhao, Z.~Wang, P.~Gao, H.~Li, D.~Wang, and X.~Li, ``Any2point: Empowering any-modality large models for efficient 3d understanding,'' in \emph{European Conference on Computer Vision}.\hskip 1em plus 0.5em minus 0.4em\relax Springer, 2024, pp. 456--473.

\bibitem{pointtransformerv3}
X.~Wu, L.~Jiang, P.-S. Wang, Z.~Liu, X.~Liu, Y.~Qiao, W.~Ouyang, T.~He, and H.~Zhao, ``Point transformer v3: Simpler faster stronger,'' in \emph{Proceedings of the IEEE/CVF conference on computer vision and pattern recognition}, 2024, pp. 4840--4851.

\bibitem{sonata}
X.~Wu, D.~DeTone, D.~Frost, T.~Shen, C.~Xie, N.~Yang, J.~Engel, R.~Newcombe, H.~Zhao, and J.~Straub, ``Sonata: Self-supervised learning of reliable point representations,'' in \emph{Proceedings of the Computer Vision and Pattern Recognition Conference}, 2025, pp. 22\,193--22\,204.

\bibitem{concerto}
Y.~Zhang, X.~Wu, Y.~Lao, C.~Wang, Z.~Tian, N.~Wang, and H.~Zhao, ``Concerto: Joint 2d-3d self-supervised learning emerges spatial representations,'' \emph{Advances in Neural Information Processing Systems}, vol.~38, pp. 69\,498--69\,522, 2025.

\bibitem{utonia}
Y.~Zhang, X.~Wu, Y.~Yang, X.~Fan, H.~Li, Y.~Zhang, Z.~Huang, N.~Wang, and H.~Zhao, ``Utonia: Toward one encoder for all point clouds,'' in \emph{ICML}, 2026.

\bibitem{bitfit}
E.~B. Zaken, S.~Ravfogel, and Y.~Goldberg, ``Bitfit: Simple parameter-efficient fine-tuning for transformer-based masked language-models,'' \emph{arXiv preprint arXiv:2106.10199}, 2021.

\bibitem{hu2021lora}
E.~J. Hu, Y.~Shen, P.~Wallis, Z.~Allen-Zhu, Y.~Li, S.~Wang, L.~Wang, and W.~Chen, ``Lora: Low-rank adaptation of large language models,'' \emph{arXiv preprint arXiv:2106.09685}, 2021.

\bibitem{fourierft}
Z.~Gao, Q.~Wang, A.~Chen, Z.~Liu, B.~Wu, L.~Chen, and J.~Li, ``Parameter-efficient fine-tuning with discrete fourier transform,'' in \emph{International Conference on Machine Learning}.\hskip 1em plus 0.5em minus 0.4em\relax PMLR, 2024, pp. 14\,884--14\,901.

\bibitem{ssf}
D.~Lian, D.~Zhou, J.~Feng, and X.~Wang, ``Scaling \& shifting your features: A new baseline for efficient model tuning,'' \emph{Advances in Neural Information Processing Systems}, vol.~35, pp. 109--123, 2022.

\bibitem{sct}
H.~H. Zhao, P.~Wang, Y.~Zhao, H.~Luo, F.~Wang, and M.~Z. Shou, ``Sct: A simple baseline for parameter-efficient fine-tuning via salient channels,'' \emph{International Journal of Computer Vision}, vol. 132, no.~3, pp. 731--749, 2024.

\bibitem{vfpt}
R.~Zeng, C.~Han, Q.~Wang, C.~Wu, T.~Geng, L.~Huangg, Y.~N. Wu, and D.~Liu, ``Visual fourier prompt tuning,'' \emph{Advances in Neural Information Processing Systems}, vol.~37, pp. 5552--5585, 2024.

\bibitem{tsne}
L.~Van~der Maaten and G.~Hinton, ``Visualizing data using t-sne.'' \emph{Journal of machine learning research}, vol.~9, no.~11, 2008.

\bibitem{fcaf3d}
D.~Rukhovich, A.~Vorontsova, and A.~Konushin, ``Fcaf3d: fully convolutional anchor-free 3d object detection,'' in \emph{European Conference on Computer Vision}.\hskip 1em plus 0.5em minus 0.4em\relax Springer, 2022, pp. 477--493.

\end{thebibliography}
\end{document}